\documentclass[sigconf,nonacm]{aamas}

\usepackage{balance} 
\usepackage{array} 
\usepackage{appendix}
\usepackage[utf8]{inputenc}
\usepackage{algorithm}
\usepackage{algorithmic}
\usepackage{xcolor}
\usepackage{amsmath}
\usepackage{makecell}
\usepackage{booktabs}
\usepackage[table]{xcolor}
\usepackage{multirow} 
\usepackage[most]{tcolorbox}   
\tcbset{enhanced,
        arc=2pt,                       
        boxrule=0.7pt,                 
        colframe=black,                
        colback=white,                 
        fonttitle=\bfseries,           
        colbacktitle=black!10,         
        coltitle=black,                
        left=3pt,right=3pt,top=3pt,bottom=3pt   
}

\doi{UEHN4980}

\makeatletter
\gdef\@copyrightpermission{
  \begin{minipage}{0.2\columnwidth}
   \href{https://creativecommons.org/licenses/by/4.0/}{\includegraphics[width=0.90\textwidth]{by}}
  \end{minipage}\hfill
  \begin{minipage}{0.8\columnwidth}
   \href{https://creativecommons.org/licenses/by/4.0/}{This work is licensed under a Creative Commons Attribution International 4.0 License.}
  \end{minipage}
  \vspace{5pt}
}
\makeatother

\setcopyright{ifaamas}
\acmConference[AAMAS '26]{Proc.\@ of the 25th International Conference
on Autonomous Agents and Multiagent Systems (AAMAS 2026)}{May 25 -- 29, 2026}
{Paphos, Cyprus}{C.~Amato, L.~Dennis, V.~Mascardi, J.~Thangarajah (eds.)}
\copyrightyear{2026}
\acmYear{2026}
\acmDOI{}
\acmPrice{}
\acmISBN{}

\title[AAMAS-2026 Formatting Instructions]{Reputation as a Solution to Cooperation Collapse \\ in LLM-based MASs}

\author{Siyue Ren}
\affiliation{
  \institution{School of Mechanical Engineering, Northwestern Polytechnical University}
  \city{Xi'an}
  \country{China}}

\author{Wanli Fu}
\affiliation{
  \institution{School of Cybersecurity, Northwestern Polytechnical University}
  \city{Xi'an}
  \country{China}}

\author{Xinkun Zou}
\affiliation{
  \institution{School of Artificial Intelligence, OPtics and ElectroNics (iOPEN), Northwestern Polytechnical University}
  \city{Xi'an}
  \country{China}}

\author{Chen Shen}
\affiliation{
  \institution{Kyushu University}
  \city{Fukuoka}
  \country{Japan}}

\author{Yi Cai}
\affiliation{
  \institution{South China University of Technology}
  \city{Guangzhou}
  \country{China}}

\author{Chen Chu}
\affiliation{
  \institution{Yunnan University of Finance and Economics}
  \city{Kunming}
  \country{China}}

\author{Zhen Wang*}
\affiliation{
  \institution{Northwestern Polytechnical University}
  \city{Xi'an}
  \country{China}}
\email{w-zhen@nwpu.edu.cn}

\author{Shuyue Hu*}
\affiliation{
  \institution{Shanghai Artificial Intelligence Laboratory}
  \city{Shanghai}
  \country{China}}
\email{hushuyue@pjlab.org.cn}

\begin{abstract}
Cooperation has long been a fundamental topic in both human society and AI systems. However, recent studies indicate that the collapse of cooperation may emerge in multi-agent systems (MASs) driven by large language models (LLMs). To address this challenge, we explore reputation systems as a remedy. We propose \emph{RepuNet}, a dynamic, dual-level reputation framework that models both agent-level reputation dynamics and system-level network evolution. Specifically, driven by direct interactions and indirect gossip, agents form reputations for both themselves and their peers, and decide whether to connect or disconnect other agents for future interactions. Through three distinct scenarios, we show that RepuNet effectively avoids cooperation collapse, promoting and sustaining cooperation in LLM-based MASs. Moreover, we find that reputation systems can give rise to rich emergent behaviors in LLM-based MASs, such as the formation of cooperative clusters, the social isolation of exploitative agents, and the preference for sharing positive gossip rather than negative ones. The GitHub repository for our project can be accessed via the following link: https://github.com/RGB-0000FF/RepuNet.
\end{abstract}

\thanks{Corresponding author: w-zhen@nwpu.edu.cn, hushuyue@pjlab.org.cn}


\newcommand{\BibTeX}{\rm B\kern-.05em{\sc i\kern-.025em b}\kern-.08em\TeX}

\begin{document}


\pagestyle{fancy}
\fancyhead{}


\maketitle 


\section{Introduction}

\label{sec: intro}

\textit{Cooperation collapse} refers to a phenomenon driven by social dilemmas, in which individual self-interest conflicts with collective welfare, resulting in harmful outcomes for the entire group~\cite{hardin1998extensions,ostrom2008tragedy,orzan2024emergent}. Such collapse occurs across diverse domains and underpins a wide array of challenges, including resource allocation~\cite{piatti2024cooperate}, climate change mitigation~\cite{dietz2020climate}, and coordination among self-driving vehicles~\cite{dafoe2021cooperative}. Addressing cooperation collapse is essential for sustaining effective collaboration in both human societies and AI systems.

Recent studies on large language model (LLM)-based agents have revealed the existence of cooperation collapse in LLM-based multi-agent systems (MASs), a system of agents that are powered by LLMs. For example, in simulated fishing allocation scenarios, Piatti et al.~\cite{piatti2024cooperate} observed that LLM-based agents tended to over-exploit shared resources, resulting in rapid depletion and long-term unsustainability. Similarly, in iterated Prisoner's Dilemma settings, agents often responded to a single defection with irreversible retaliation, leading to cascading mutual defection and the breakdown of cooperation~\cite{akata2025playing,wang2024large,gandhi2023strategic}. While these studies have documented the widespread cooperation collapse, effective solutions remain elusive.

\begin{figure*}[htbp]
  \centering
  \includegraphics[width=\textwidth]{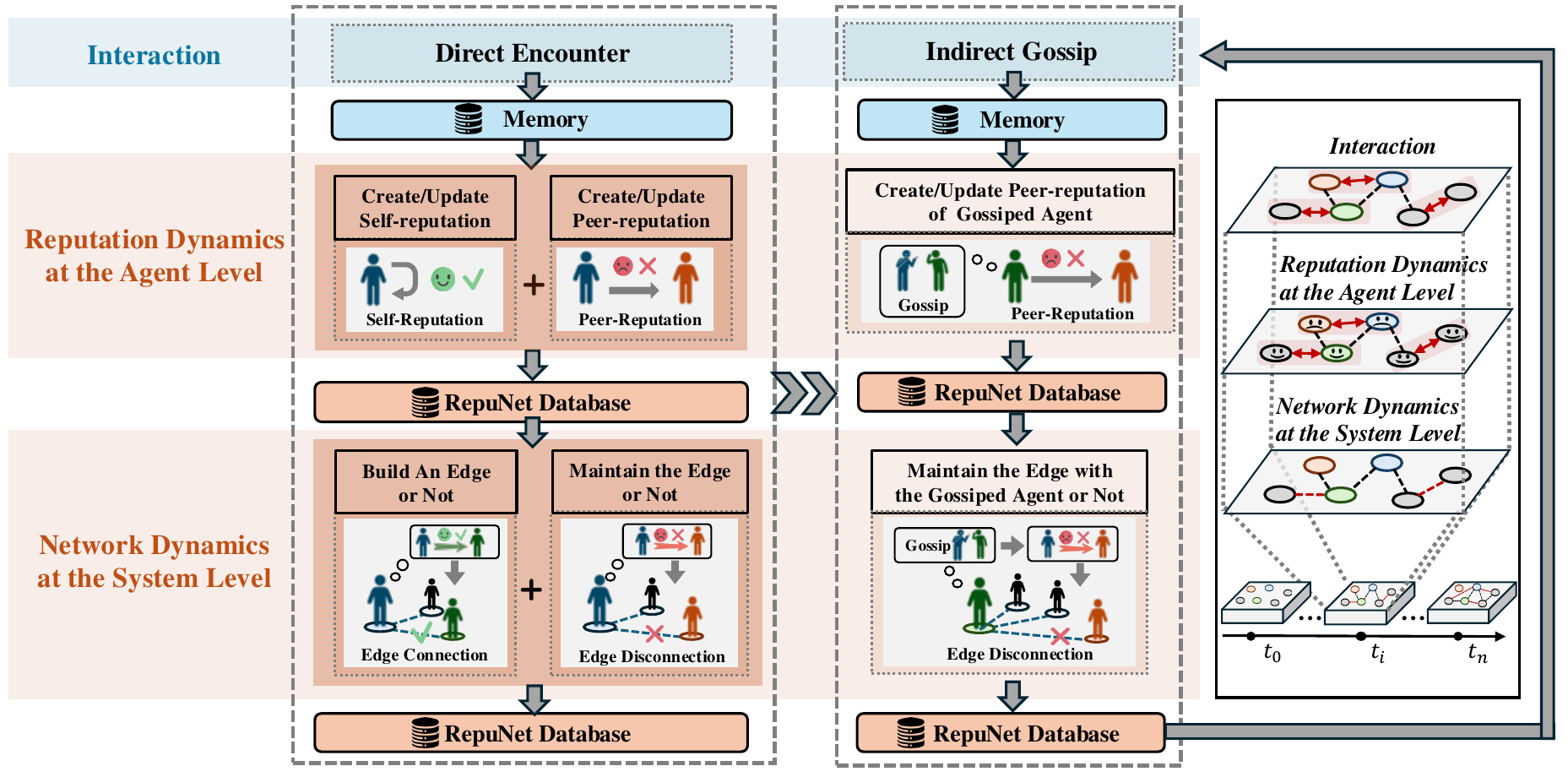}
  \caption{RepuNet:  A dynamic reputation system aimed at sustaining cooperation and preventing cooperation collapse in LLM-based MASs. Agents interact within networks through both direct encounters and indirect gossip, shaping their self-reputation and peer-reputation. At the system level, agents decide whether to form or maintain network connections (i.e. edges) based on these reputations. The evolving reputations and network structures, stored in RepuNet databases, continuously guide agents' behaviors, influencing their future interactions.}
  \label{fig:reputation system}
\end{figure*}

This paper explores the use of reputation systems as a remedy. Reputation is a set of collective beliefs or evaluations about individuals based on their past behavior, enabling agents to form expectations about others~\cite{bromley1993reputation,sperber2012moral,mu2024multi,santos2019outcome}. As an emergent social signal, it encodes perception of trustworthiness, reliability, and integrity.
A reputation system operationalizes this signal by systematically collecting, aggregating, and disseminating reputational information across agents.
Proven effective in both human societies and traditional MASs, reputation systems offer a fesible solution to fostering cooperation even among strangers, evoking social norms, and serving as lightweight, scalable substitutes for direct monitoring or formal enforcement~\cite{suzuki2005reputation,barclay2012harnessing,giardini2022four,xu2019cooperation}.

In this research, we extend reputation mechanisms to LLM-based MASs, and demonstrate that our approach effectively prevents cooperation collapse. Crucial to our study will be how to \emph{operationalize} reputation mechanisms for LLM-based MASs. To this end, we introduce \textbf{RepuNet}, the first \emph{operational} reputation system tailored for LLM-based MASs, in which every interaction is explicitly embedded within a network structure. RepuNet operates on two levels: agent-level reputation dynamics and system-level network evolution. Specifically, at the agent level, echoing real‐world reputation dynamics~\cite{sabater2004trust,granatyr2015trust,giardini2022four}, RepuNet enables agents shape initial self- and peer-reputations after their first encounter and iteratively update them after each subsequent interaction. Besides, gossip has been a key mechanism for spreading reputational information across networks throughout human evolution. Inspired by this, agents evaluate each direct encounter and, when (dis)satisfied, broadcast that information, updating the target’s reputation. In RepuNet, we explicitly couple reputation updates to topology changes: once a reputation is revised, it drives network rewiring. At the system level, agents strengthen ties to high-reputation partners and sever links to disreputable ones, reflecting the human tendency to seek trustworthy collaborators and abandon exploitative relationships. This network rewiring reshapes future encounter probabilities and fosters long-term cooperation among reputable agents. We implement RepuNet by prompting LLM-based agents, rather than relying on any rule-based mechanisms or mathematical formulations. Both reputation dynamics and network evolution in RepuNet are autonomously determined by the agents themselves, without any human intervention.
An overview of our reputation system is provided in Figure \ref{fig:reputation system}.

In the experiment, we consider three scenarios that progress from classical to real-world dilemmas: (i) \textit{the Prisoner’s Dilemma}, a fundamental social dilemma in game theory~\cite{axelrod1981evolution}; (ii) \textit{voluntary participation}, adapted from human field experiments~\cite{yoeli2013powering}, modeling collective resource-sharing; and (iii) \textit{trading investment}, based on an economic trust game~\cite{berg1995trust}, capturing sequential investment dilemmas. Within LLM-based MASs, our experiments replicate the phenomenon of cooperation collapse previously observed in human studies. Specifically, agents acting without RepuNet prioritize individual gain, causing widespread exploitation and eventual cooperation breakdown. In contrast, with RepuNet, agents are incentivized to maintain cooperation, favoring collective welfare and preventing collapse. Across all three scenarios, we observed consistent outcomes using similar prompts with varying task descriptions. This demonstrates robustness to wording variations, ensuring our results are not artifacts of a single idiosyncratic prompt or scenario.
Additionally, our experiments reveal rich emergent behaviors: (i) cooperative agents with high reputations self-organize into persistent, tightly connected clusters; (ii) these clusters selectively isolate low-reputation agents who exhibit defective behavior; (iii) unlike humans, LLM-based agents prefer sharing positive gossip rather than negative; and (iv) without RepuNet, dishonesty leads to network collapse. An ablation study further confirms the effectiveness of each RepuNet component: removing any single module reduces cooperation rates and may trigger complete collapse. The experiment’s GitHub repository is accessible via the following link: https://github.com/RGB-0000FF/RepuNet.

In summary, our key contributions are as follows:
\begin{enumerate}
    \item We replicate the phenomenon of cooperation collapse within LLM-based MASs, previously observed in human studies~\cite{yoeli2013powering}, confirming its widespread prevalence and emphasizing the critical need to address this challenge.

    \item We introduce \emph{RepuNet}, the first \emph{operational} system to generalize reputation mechanisms to LLM-based MASs. It not only establishes reputation dynamics through direct social interactions and indirect gossip but also drives network evolution, thereby effectively preventing cooperation collapse.

    \item We show \emph{RepuNet} exhibits various emergent phenomena, such as cooperative clustering, the isolation of exploitative agents, and a bias toward positive gossip, demonstrating the crucial role of reputation systems in catalyzing complex collective behaviors in LLM-based MASs.

\end{enumerate} 

\section{Background}
\label{sec: related work}
In this section, we define and formalize the cooperation problem. Next, we review recent studies identifying cooperation collapse in LLM-based MASs. We then discuss prior research on how reputation and social networks facilitate cooperation in traditional MASs.

\subsection{Cooperation Problem}
In this study, we investigate the cooperation problem within the classic context of social dilemmas~\cite{nowak2006five,conitzer2023foundations}, where individual interests conflict with collective welfare.
It is important to distinguish this setting from a separate line of research on LLM-based multi-agent coordination~\cite{chen2023agentverse,hong2024metagpt,chen2024blockagents}. Although some prior works use the terms cooperation and coordination interchangeably, coordination typically refers to scenarios without conflicting interests, where agents work together toward a shared objective.

A social dilemma, in its elementary form, can be exemplified by the two-player Prisoner's Dilemma, in which each agent independently decides either to cooperate (C) or defect (D). Payoffs for each player are summarized in the following matrix:

\[
\begin{array}{c|cc}
& \text{C} & \text{ D} \\ 
\hline
\text{C} & (R, R) & (S, T) \\
\text{D} & (T, S) & (P, P)
\end{array}
\]

A social dilemma occurs under the payoff conditions $T > R > P > S $ and $2R > T + S$. Although mutual cooperation maximizes collective benefits, individual rationality incentivizes defection, leading to worse outcomes for everyone---a phenomenon known as \emph{cooperation collapse}. While illustrated here as a binary-choice, two-player scenario, social dilemma can be extended to multi-player and multi-choice situations, such as the Public Goods Game.

\subsection{Cooperation Problem in LLM-based MASs}
Recent studies have explicitly identified the presence of cooperation collapse in MASs~\cite{piatti2024cooperate,mozikov2024eai,mao2023alympics,akata2025playing,ren2024emergence}. Mozikov et al.~\shortcite{mozikov2024eai} examined how simulated emotions affect LLM-based agents' behavior in social dilemmas, finding that even positive emotions can unpredictably reduce cooperation among LLM-based agents. Piatti et al.~\shortcite{piatti2024cooperate} proposed a simulation platform to study resource-sharing scenarios, revealing that agents frequently over-exploited shared natural resources, thereby failing to maintain sustainable cooperation and exacerbating the collapse of cooperation.  Mao et al.~\shortcite{mao2023alympics} showed that, in a multi-round water-auction game, a lower initial resource-supply ratio provokes LLM-based agents to bid more aggressively, reducing overall survival rates. Akata et al.~\shortcite{akata2025playing} found that, in iterated two-strategy games, these LLM-based agents often permanently defect following a single betrayal, hindering long-term cooperation. While these studies highlight the problem, they primarily focus on analyzing LLM-based agents' performance rather than proposing solutions to avoid cooperation collapse, which remains a pressing open challenge in LLM-based MASs.

\subsection{Reputation for Traditional Multi-agent Systems} 
Indirect and network reciprocity are key mechanisms for the evolution of cooperation~\cite{nowak2006five}. Reputation underpins indirect reciprocity by allowing individuals to condition their behavior on others’ reputations, enabling higher cooperation than reputation-blind strategies~\cite{nowak2005evolution}. Network reciprocity, on the other hand, promotes cooperation by restricting interactions to local neighborhoods, in contrast to well-mixed populations where players interact globally~\cite{nowak1992evolutionary}. Dynamic networks enhance this effect by enabling individuals to rewire ties, fostering assortative interactions among cooperators~\cite{rand2011dynamic,wang2012cooperation}. Some works implement this interplay that specify how reputations are updated~\cite{gross2019rise}, how networks evolve~\cite{fu2008reputation}, and how agents behave~\cite{anastassacos2021cooperation}. Although they cannot provide a direct solution due to their historical inability to leverage the strengths of LLMs, they provide us with valuable insights for addressing cooperation collapse in LLM-based MASs.

\subsection{Social networks in LLM-based MASs}
Social networks profoundly influence LLM-based MASs by shaping agent interactions, information dissemination, and behavioral coordination~\cite{takacs2021networks}. Recent studies have leveraged LLM-based agents to simulate complex dynamics within social networks, aiming to explore emergent human-like behaviors~\cite{gao2023s3,williams2023epidemic,yang2024oasis,chuang2024simulating}. For instance, Gao et al.~\cite{gao2023s3} introduced the S3 framework, which employs LLM-based agents to replicate nuanced individual and collective human behaviors---including emotions, attitudes, and interactions---within real-world social networks. Yang et al.~\cite{yang2024oasis} developed the OASIS framework, a simulated social media network capable of scaling to millions of agents, facilitating extensive investigations into collective phenomena such as viral information diffusion, echo chamber formation, and group polarization dynamics. Although these studies highlight significant progress in LLM-based MASs for modeling emergent group behaviors, the fundamental challenge of sustaining cooperation within evolving social networks remains underexplored.


\section{RepuNet: a Reputation System for LLM-based Agent Societies }
\label{sec: method}
In this section, we formalize our reputation-based MAS. After an overview of the networked system, we introduce RepuNet from two perspectives: (i) agent-level reputation dynamics, and (ii) system-level network evolution. Due to space limitations, detailed prompts for RepuNet operations are provided in Appendix~B (which is only available in the arXiv version of the paper).

\subsection{Formalization: Networked Multi-Agent System and Reputation}
\label{sec:3.2}
In a MAS, interactions between agents can be represented by a network, which determines the likelihood of whom they meet and interact with. 
 Let
 an ordered pair $G=(V, E)$ denote a directed graph, where $V$ is the set of vertices representing the agents in the system, and $E \subseteq \{(i,j)|i,j\in V^2$ and $i\neq j\}$ is the set of directed edges, each of which is an ordered pair of vertices. A directed edge from vertex $i$ to vertex $j$ indicates that agent $j$ is reachable from agent $i$, meaning that agent $j$ is viewed as a potential partner that agent $i$ is willing to interact in future interactions.

In our system, reputation arises from information exchange within the embedded network. Specifically, two types of reputations can emerge in our reputation-based MAS: (i) self-reputation, which reflects an agent’s self-perception of how others evaluate it based on past interactions~\cite{giardini2022four}, and (ii) peer-reputation, which captures an agent’s evaluation of other’s behaviors in previous interactions~\cite{bromley1993reputation,takacs2021networks}. Regardless of the types, formally, reputation in our system can be represented as a standardized quintuple: $r = \langle a, s, o, c, \mu \rangle$. Here, $a$ denotes the basic information of the agent being evaluated (i.e., the evaluated target), such as its ID and name. Since reputation is context-dependent (e.g., one may be a good parent but not a good colleague)~\cite{sabater2001regret}, $s$ and $o$ denote the scenario and role of the evaluated target, respectively. $c$ represents the reputation in natural language, e.g., ``James is too self-centered, often prioritizing his interests over the community's needs''; $\mu$ quantifies the reputation described in $c$, ranging from -1 to 1, with higher scores indicating a better reputation. We consider that for each agent, the  self-reputation and peer-reputation are stored in its RepuNet database, denoted as $\mathcal{R}$. Note that an agent’s reputation always refers to the most recently stored value in the database, as reputations are dynamically updated.

To allow RepuNet to be integrated with other MASs involving agent interactions, such as Generative Agents~\cite{park2023generative} and ProAgents~\cite{zhang2024proagent}, we consider it as an event-driven system that responds to agent encounters. Driven by direct encounters and indirect gossip, our reputation-based MAS dynamically update reputations in response to new interactions, allowing evaluations to adapt to evolving agent behaviors. We model these dynamics on two levels: reputation dynamics at the agent level and network dynamics at the system level, which will be discussed in following sections.

\subsection{Reputation Dynamics at the Agent Level}
This section explores how reputations are shaped and updated through direct encounters and indirect gossip~\cite{mui2002notions,wu2016reputation,giardini2022four}.

\subsubsection{Reputation Driven by Direct Encounters}

In the real world, reputation is often shaped by direct social encounters, such as when a seller provides quality products~\cite{sabater2004trust,granatyr2015trust}. Reputation is not fixed or unchanging; rather, it evolves over time based on new encounters and the cumulative evaluations of one's past behavior, that is, one's previous reputation~\cite{giardini2022four}. Inspired by these, in our system, agents can form both self-reputation and peer-reputation based on their first social encounter. Over time, as agents meet and interact again, they update these reputations by considering both previous reputations and the impressions formed during the latest interaction. The process of reputation formation and updating is implemented through prompting LLMs, rather than rule-based mechanisms or mathematical formulations.

Specifically, let $m_{ij}(t)$ represent the direct encounter between agent $i$ and $j$ at time $t$. After they first encounter, for instance, the agent $i$ generates a peer-reputation for the agent $j$, denoted as $r_{i\rightarrow j}(t)$. As time evolves, on subsequent encounters, agent $i$  updates agent $j$'s reputation based on the new encounter $m_{ij}(t+t')$ and previous peer-reputation $r_{i\rightarrow j}(t)$. We represent this process by a LLM-based operation $r_{i\rightarrow j}(t+t') \leftarrow \texttt{ShapeRepuPeer}(m_{ij}(t+t'), r_{i\rightarrow j}(t), \text{if $r_{i\rightarrow j}(t)\neq\phi$ })$. By shaping peer-reputation, we expect agents are able to select partners rather than relying on random encounters, avoiding free-riders and fostering cooperation in LLM-based MASs.

In addition to peer-reputation, agents can also generate self-reputations to reflect their self-perceptions. To ensure alignment with their personalities, LLM-based agents are prompted to generate self-reputations based on both their agent descriptions $\mathcal{D}$ and the direct encounter $m_{ij}(t)$. Similar to shaping peer-reputation, as time evolves, once agent $i$ encounters another agent $x$, not necessarily the previous agent $j$, we instruct agent $i$ to update its self-reputation based on the new encounter $m_{ix}(t+t')$ and its previous self-reputation $r_i(t)$. The process can be represented by the LLM-based operation $r_i(t+t')\leftarrow\texttt{ShapeRepuSelf}(\mathcal{D}_i,m_{ix}(t+t'), r_i(t),\text{if $r_i(t)\neq\phi$ })$. By shaping self-reputation, we aim to help agents maintain appropriate behaviors to uphold good reputations for cooperation. 

\subsubsection{Reputation Driven by Indirect Gossip}
Gossip has been a part of human interactions throughout evolutionary history, from early hunter-gatherer societies to today’s social media-driven culture~\cite{lange2014reward,wu2016gossip}. It involves the exchange of positive or negative information between a gossiper and a listener about an \emph{absent} individual, making it an effective way to shape and spread reputations~\cite{foster2004research,giardini2012gossip}. In human societies, people have a strong tendency to gossip about others when they are (dis)satisfied with interactions, using it to influence others' reputations~\cite{foster2004research,wu2016reputation}. Inspired by this, our system enables each agent to evaluate its satisfaction with others’ behavior based on their innate attributes described in agent descriptions and then decide whether to gossip after each interaction. Due to the lack of space, details of the gossip process and related prompts are provided in Appendix~A and~B.

When gossip occurs, a listener generates a peer-reputation for the agent being gossiped based on the gossip information (i.e., what the gossiper conveys).  Let $\theta_{y}(t)$ denote the gossip information about the agent being gossiped $y$ at the time $t$. The listener $l$ shape a peer-reputation of agent $y$, denoted as $r_{l\rightarrow y}(t)$. As time evolves, when agent $l$ hears new gossip about agent $y$ at time $t+t'$, we instruct agent $l$ to update peer-reputation by considering both the latest gossip information $\theta_y(t+t')$ and the previous peer-reputation $r_{l\rightarrow y}(t)$. The LLM-based operation can be represented by $r_{l\rightarrow y}(t+t')\leftarrow\texttt{ShapeRepuGossip}(\theta_y(t+t'), r_{l\rightarrow y}(t), \text{if $r_{l\rightarrow y}(t)\neq\phi$})$. Through gossip, we aim to empower agents to exchange reputations of unknown agents, helping them assess potential partners and reduce defection risks.

\subsection{Network Dynamics at the System Level}
This section focuses on the system level, examining how networks evolve. Once an agent shapes peer-reputations, it is natural to consider whether to build or maintain edges with them for future interactions based on their reputations. As RepuNet dynamics are event-driven, we analyze network evolution via direct encounters and gossip.

\begin{figure*}[htbp]
  \centering
  \includegraphics[width=0.95\textwidth]{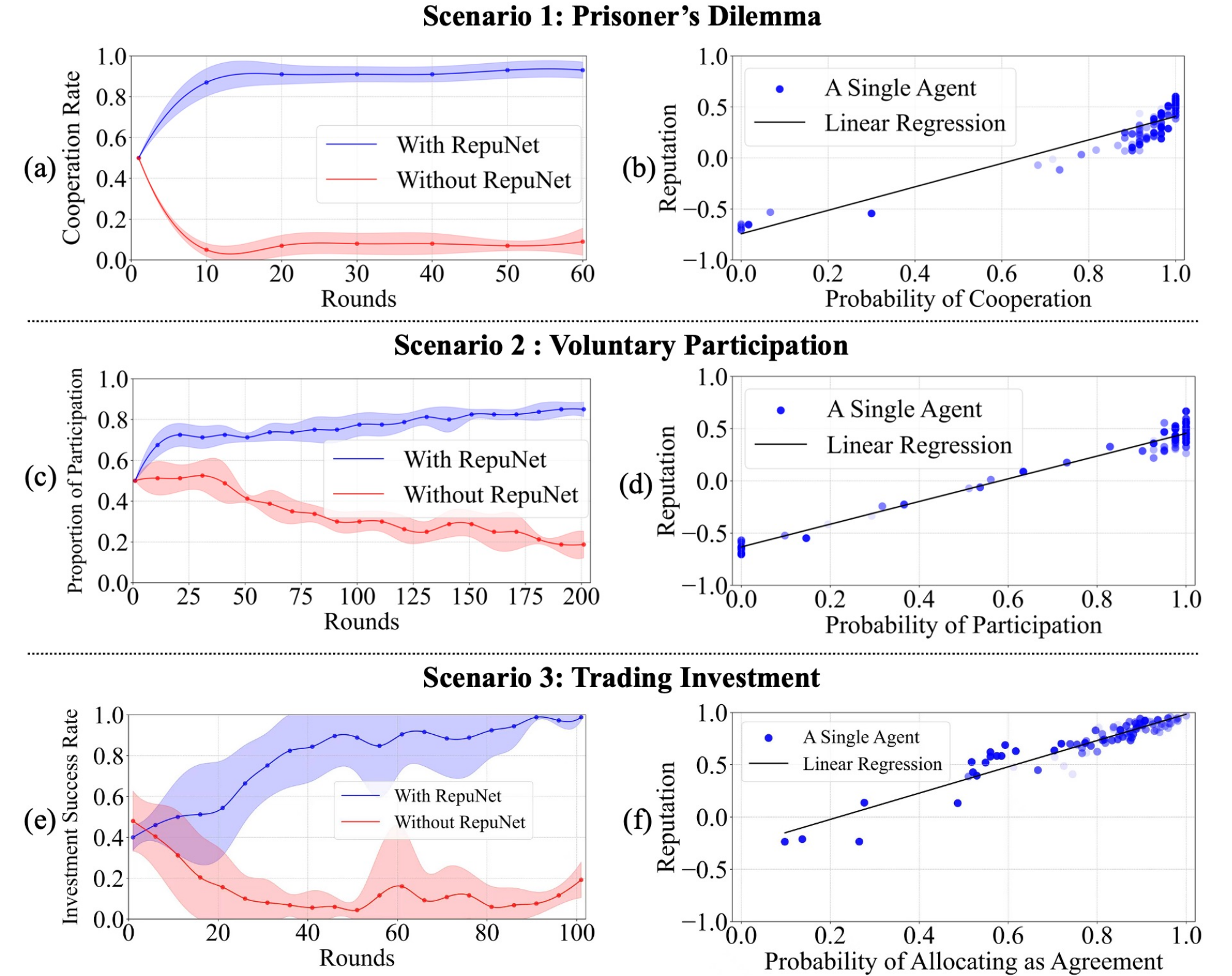}
  \caption{Experimental results across three scenarios. Panels (a), (c) and (e) illustrate the trends of agent cooperation, participation, and investment success rates, respectively. Solid lines represent average rates, and shaded areas indicate error margins. Panels (b), (d) and (f) show statistically significant correlations (all $p < 0.001$) between agent reputation and behavior, illustrated by linear regression. Each data point represents an agent's average behavior over the last ten rounds, across five experimental runs (indicated by different shades of blue).}
  \label{fig:results-1}
\end{figure*}

\subsubsection{Network Driven by Direct Interactions}
In human society, social networks determine the likelihood of individuals meeting, interacting, and exchanging information~\cite{takacs2021networks}. Individuals seeking new partners prefer to build relations with reputable ones and avoid exploitation by unilaterally ending unfavorable relationships~\cite{bshary2005punishment,du2011partner}. Inspired by these, we instruct each LLM-based agent to decide whether to build or maintain an edge with another to enable future interactions. Specifically, let $\mathcal{D}_i$ represent agent $i$'s agent description, and  $m_{ij}(t)$ denote the direct encounter between agent $i$ and $j$ at time $t$. Once agent $i$ shapes peer-reputation $r_{i\rightarrow j}(t)$ of the agent $j$, it considers whether to build or maintain the edge with the agent $j$ for future interactions. We represent this LLM-based operation by $w_{i\rightarrow j}(t)\in\{\text{``Y'',N''}\} \leftarrow \text{InteractEdgeShape}(\mathcal{D}_i, m_{ij}(t), r_{i\rightarrow j}(t))$. Once the edge is formed (i.e., $w_{i\rightarrow j}(t)=\text{``Y''}$), the agent $i$ records an ordered pair $(i,j)$ in its RepuNet database $\mathcal{R}_i$. As the network evolves, we expect agents with high reputations form more edges and dense clusters, fostering relationships that sustain cooperation. Meanwhile, low-reputation agents form fewer edges and may become isolated from cooperative ones.

\subsubsection{Network Driven by Indirect Gossip}
Through indirect gossip, agents (as listeners) can shape the reputations of agents being gossiped they have never met. However, they cannot build new edges with these agents, as no direct encounter has occurred. If the listener and the agent being gossiped have previously interacted, the listener can reconsider whether to maintain the edge after updating the peer-reputation of the agent being gossiped. Specifically, let $D_l$ represent the listener $l$'s agent description,  $\theta_{y}(t)$ represent the gossip information about the agent $y$ at time $t$, and $r_{l\rightarrow y}(t)$ denote the listener $l$ shapes a peer-reputation of agent $y$. We instruct LLM-based agents to consider whether to maintain an edge with the agent being gossiped based on the aforementioned information, which can be formally represented by the LLM-based operation $w_{l\rightarrow y}(t)\in\{\text{``Y'',N''}\} \leftarrow \text{GossipEdgeShape}(\mathcal{D}_l, \theta_{y}(t), r_{l\rightarrow y}(t))$. Once the edge is disconnect (i.e., $w_{l\rightarrow y}(t)=\text{``N''}$), agent $l$ removes the ordered pair $(l,y)$ from its RepuNet database $R_l$, thereby causing the network to evolve at time $t$. By leveraging gossip-driven reputation updates, we aim to ensure that agents can promptly update peer reputations, which prevents the stagnation of reputation updates due to a lack of further encounters.


\section{Experiment}
\label{sec:experiment}
Our experiments aim to answer three key questions across three scenarios: (i) Does our RepuNet effectively avoid  cooperation collapse? (ii) How does our RepuNet prevent the occurrence of the collapse? (iii) How effectively does each component of our RepuNet contribute to sustaining long-term cooperation? We first outline the experimental settings, then address the first two research questions, and finally examine the last question through an ablation study.

\subsection{Experimental Settings}
\label{sec:4.1Voluntary Participation}

\paragraph{\textbf{Initialization}} We conducted experiments with 20 LLM-based agents, initially placed as isolated nodes without connections and varying in initial preferences: some were prosocial, prioritizing collective welfare, while others were self-interested, favoring individual gains. They interacted pairwise in three distinct scenarios, gradually forming network edges through random or reputation-based partner selection. Experiments ended once cooperation rates stabilized, consequently, rounds varied by scenario. For computational efficiency, we used GPT-4o mini, repeating each scenario five times for reliability. We additionally conducted experiments with three distinct LLMs to verify RepuNet's robustness. The corresponding details are provided in Appendixes B and C, which are included exclusively in the arXiv version. Codes are available at: https://github.com/RGB-0000FF/RepuNet.

\paragraph{\textbf{Controlled Settings}}
To evaluate RepuNet, we also conducted controlled settings using a standard LLM-based MAS architecture~\cite{xi2025rise,mou2024individual}, comprising three modules: \emph{profile} (defining roles and preferences), \emph{memory} (storing perceived information for behavioral consistency), and \emph{action} (enabling interaction with the environment and other agents). The key distinction is that agents without RepuNet interact only through the evolved network, lacking the capability to evaluate peers, form reputations, or spread peer reputations through gossip.

\paragraph{\textbf{Scenario 1: The Prisoner's Dilemma}}
We first consider the Prisoner's Dilemma, a classic social dilemma from game theory. In this scenario, agents are paired and independently choose either to cooperate (C) or defect (D), resulting in the following payoffs: mutual cooperation yields (3,3); mutual defection yields (1,1); if one defects while the other cooperates, the defector receives 5 and the cooperator receives 0. The dilemma is that while mutual cooperation maximizes collective benefits, rational self-interest incentivizes defection, as defection consistently offers a higher individual payoff regardless of the other agent's action.

\paragraph{\textbf{Scenario 2: Voluntary Participation}}
We adapted the scenario presented by Erez et al.~\cite{yoeli2013powering}, which models a public goods dilemma in the real world. Agents decide whether to participate or not in a voluntary program to reduce energy consumption. The dilemma is that while participation lowers overall energy demand and benefits the collective, agents are often incentivized to not-participate for self-interest due to the inconvenience of reduced energy use. 
Each agent independently decided whether to participate and could revise its decision every five rounds of interactions (i.e., communications) with others.

\paragraph{\textbf{Scenario 3: Trading Investment}}
Inspired by the trust game in an investment setting~\cite{berg1995trust}, we designed a sequential investment dilemma between an investor and a trustee. The trustee proposes an allocation; if accepted, the investor invests funds, which double for the trustee to allocate as agreed or deviate. Both agents then evaluate each other. The dilemma arises since adherence promotes cooperation, yet short-term gains incentivize deviation. Each agent started with 10 units, randomly assigned as investor or trustee, and paired either randomly or by choosing partners based on reputation.

\begin{figure*}[htbp]
  \centering
  \includegraphics[width=\textwidth]{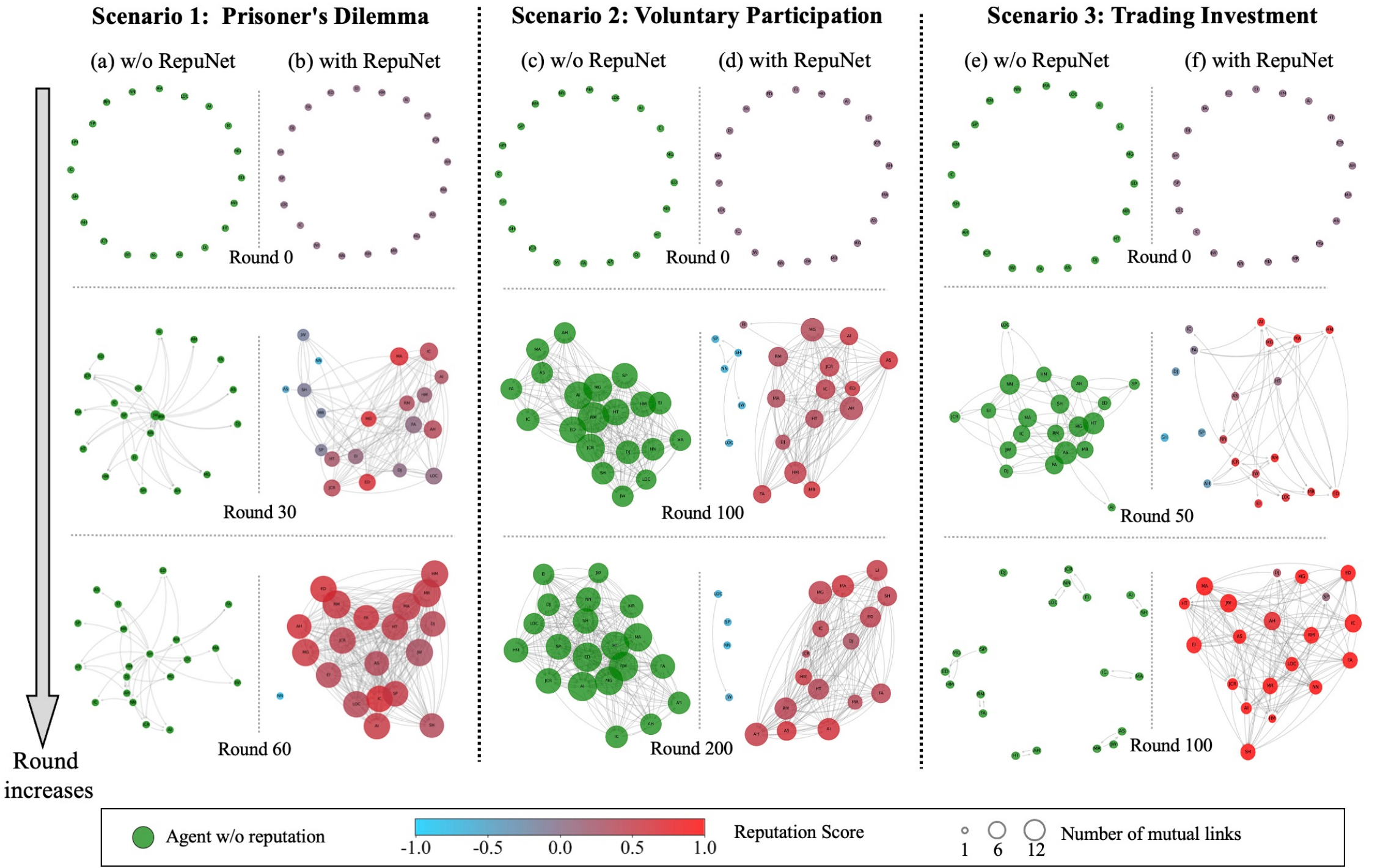}
  \caption{Case studies of network dynamics at the system level. Nodes represent agents, colored from blue (low reputation) to red (high reputation); green nodes indicate agents without RepuNet. Node size reflects the number of mutual connections. Panels (a), (c), and (e) show the network dynamics of agents without RepuNet, whereas agents with RepuNet form dense clusters and isolate low-reputation peers, as shown in panels (b), (d), and (f).}
  \label{fig:network}
\end{figure*}

\subsection{Emergent Phenomena from Three Scenarios}
\label{sec:4.2}

In this section, we evaluate RepuNet's effectiveness in preventing  cooperation collapse at two levels. At the agent level, we assess whether agents sustain cooperative behaviors to maintain high reputations. At the system level, we examine whether RepuNet drives network evolution through effective partner selection. 

\paragraph{\textbf{RepuNet sustains cooperation and effectively prevents cooperation collapse.}}
A key finding across all scenarios is that agents equipped with RepuNet consistently exhibit cooperative behaviors throughout all five independent runs. Specifically, in Scenario 1, agents with RepuNet maintain higher cooperation rates, preventing cooperation collapse (Figure~\ref{fig:results-1}a). Similarly, in Scenario 2, agents consistently participate to support public goods (Figure~\ref{fig:results-1}c). And agents in Scenario 3 also prioritize long-term cooperation over short-term self-interest, gradually reaching nearly full adherence to collective agreements (Figure~\ref{fig:results-1}e). In contrast, cooperation consistently declines in scenarios without RepuNet, ultimately resulting in cooperation collapse. In summary, RepuNet effectively incentivizes sustained cooperation, encouraging most agents to prioritize collective interests---even at the expense of personal convenience---to avoid cooperation collapse.

\begin{table*}[!h]
  \centering
  \fontsize{10pt}{12pt}\selectfont
  \begin{tabular}{>{\centering\arraybackslash}m{0.5\columnwidth}|>{\centering\arraybackslash}m{0.3\columnwidth}|>{\centering\arraybackslash}m{0.3\columnwidth}|>{\centering\arraybackslash}m{0.3\columnwidth}}  
    \toprule
    \textbf{Treatment} & \textbf{Scenario 1} & \textbf{Scenario 2} & \textbf{Scenario 3} \\
    \midrule
    with \emph{RepuNet}  & \textbf{0.93}($\pm 0.04$)   & \textbf{0.85}($\pm 0.03$)  & \textbf{0.98}($\pm 0.02$)             \\ 
    w/o Gossip    & 0.90($\pm 0.04$)   & 0.81 ($\pm 0.04$) & 0.96($\pm 0.01$)
                        \\ 
        w/o Reputation & 0.46($\pm 0.21$)  & 0.29 ($\pm 0.07$) & 0.26($\pm 0.34$)
                      \\ 
         w/o \emph{RepuNet} & 0.09 ($\pm 0.07$) & 0.19 ($\pm 0.06$) & 0.17($\pm0.08$)                        \\
    \bottomrule
  \end{tabular}
  \caption{Ablation study confirms RepuNet's effectiveness, achieving the highest cooperation rates (93\% in Scenario 1; 85\% in Scenario 2; 98\% in Scenario 3). Results are averaged over the last five rounds across five runs.}
  \label{tab:ablation study}
\end{table*}

\paragraph{\textbf{Cooperative behavior fosters good reputations, sustaining long-term cooperation through a positive feedback loop in LLM-based MASs.}}
As shown in Figure~\ref{fig:results-1}(b), (d) and (f), we observe a strong positive correlation between cooperative actions and reputations. In Scenario 1, agents that frequently cooperate achieve higher reputations. Similarly, agents in Scenario 2 who consistently participate in voluntary programs to enhance public goods obtain significantly higher average reputations. Likewise, agents in Scenario 3 who prioritize collective agreements and long-term cooperation over short-term self-interest gain higher reputations. These results indicate that cooperative behavior directly strengthens an agent's value, motivating agents to sustain their positive reputations through continued cooperation and thus reinforcing a self-sustaining cooperative cycle.

\paragraph{\textbf{RepuNet drives network evolution by clustering cooperative agents with high reputations and isolating defectors with low reputations.}} Figure~\ref{fig:network} presents case studies across three scenarios, showing snapshots from one of five experimental runs to illustrate how network structures evolve over time. Initially, agents in all three scenarios start unconnected. In Scenario 1, agents without RepuNet form only sparse and loose connections (Figure~\ref{fig:network}a), whereas agents with RepuNet gradually develop selective links, clustering high-reputation cooperators and isolating defectors (Figure~\ref{fig:network}b). A similar pattern emerges in Scenario 2 (Figure~\ref{fig:network}d). In contrast, agents without RepuNet in Scenario 2 form indiscriminate connections, interacting broadly without selectivity (Figure~\ref{fig:network}c). In Scenario 3, RepuNet enables agents to progressively form cohesive clusters of cooperative, reputable individuals, eventually integrating all high-reputation agents (Figure~\ref{fig:network}f). This emergent structure---where cooperators cluster and defectors become isolated---prevents exploitation, stabilizes cooperation, and effectively averts the collapse of cooperative behavior.

\begin{figure}[htbp]
  \centering
  \includegraphics[width=0.45\textwidth]{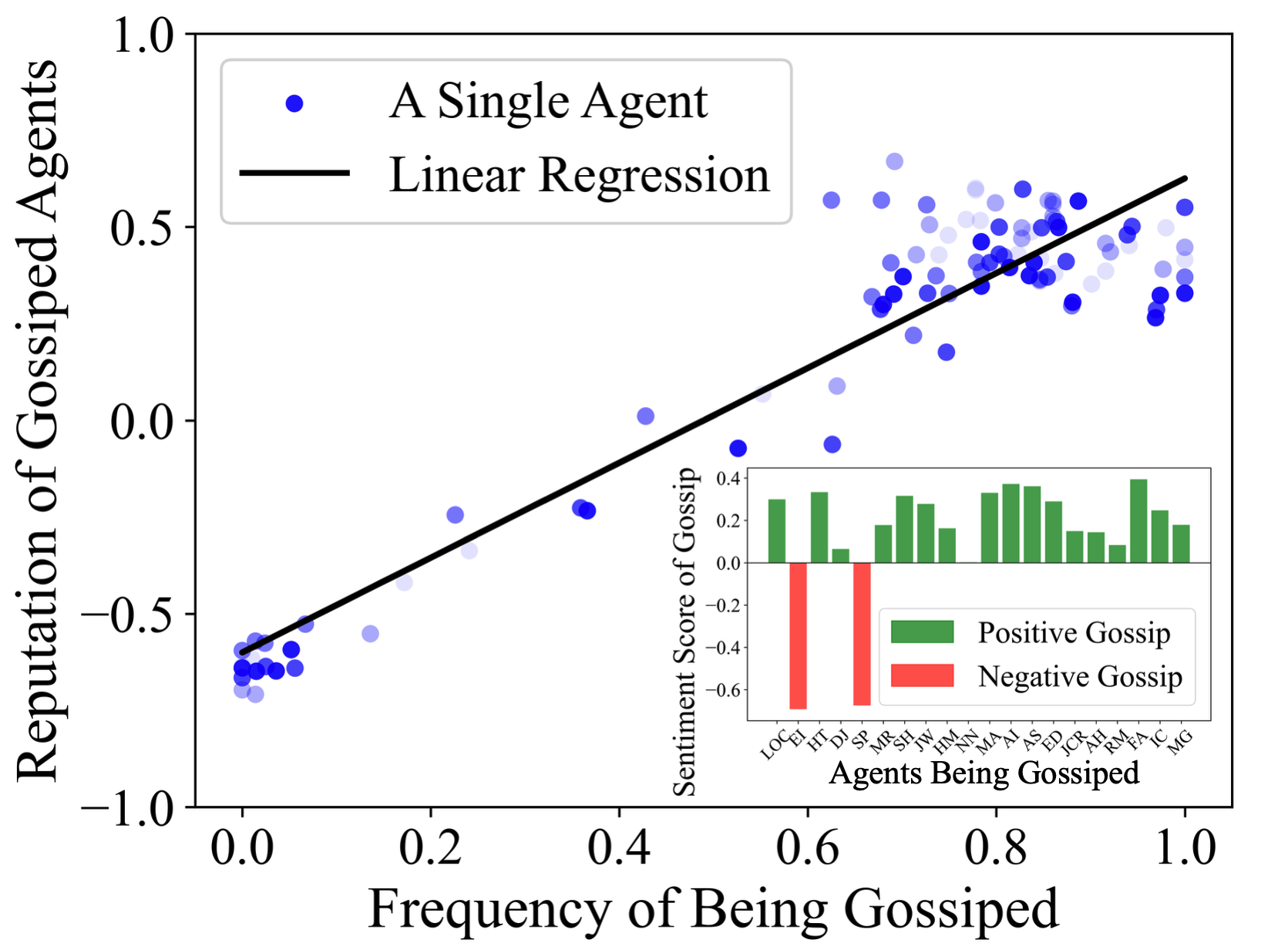}
  \caption{Correlation between gossip frequency and reputation (Scenario~2). Agents gossiped about more frequently have higher reputations, as $90\%$ of gossip is positive. The regression line shows a significant positive trend ($p<0.002$). Each shade of blue represents one of five experimental runs (average across 20 agents).}
  \label{fig:results-3}
\end{figure}

\vspace{-0.5cm}
\paragraph{More finding in Scenario 2.}
\textit{\textbf{Unlike humans~\cite{foster2004research}, LLM-based agents predominantly share positive gossip rather than negative.}} Figure~\ref{fig:results-3} highlights this with two key analyses. First, our correlation study shows that frequently discussed agents tend to have higher reputations, contrasting typical human social dynamics~\cite{foster2004research}. To investigate further, we conducted sentiment analysis using \texttt{twitter-roberta-base-sentiment}, classifying gossip as positive, neutral, or negative with corresponding confidence scores. Sentiments were mapped to numerical values ($+1$ for positive, $0$ for neutral, $-1$ for negative), weighted by confidence and averaged. The analysis revealed that approximately $90\%$ of gossip is positive, exemplified by statements such as: ``I want to gossip about Isabella... her perspective on the program as an empowering opportunity for our community is just remarkable.''

\paragraph{More finding in Scenario 3.}
\textit{\textbf{Dishonesty leads to network collapse.}} Figure~\ref{fig:network}(c) shows that, without RepuNet, networks in Scenario~3 initially form loose clusters but ultimately disintegrate. Agents prioritize short-term self-interest, exploiting benefits rather than adhering to agreed distributions. This absence of self-regulation results in widespread defection from cooperative agreements, triggering systemic instability and exemplifying cooperation collapse.

\vspace{-0.3cm}
\subsection{Ablation Study on the Reputation System}
\label{sec:4.3 ablation study}
In networked MASs, we conducted an ablation study to assess the effectiveness of RepuNet's reputation and gossip modules. Specifically, we compared four conditions: (i) \textbf{with RepuNet}: the full model, including both reputation generation and gossip interactions; (ii) \textbf{without Gossip}: agents generate and update reputations solely through direct encounters, without gossip; (iii) \textbf{without Reputation}: agents directly interact and gossip about others, but no reputations are generated or updated; and (iv) \textbf{without RepuNet}: both gossip and reputation mechanisms are completely removed, leaving agents to interact only through direct encounters within network. 

As shown in Table~\ref{tab:ablation study}, performance was averaged over the final 5 rounds in each scenario. RepuNet achieved the highest cooperation rates in all scenarios, confirming its effectiveness in preventing cooperation collapse. Removing gossip led to only a slight performance drop, suggesting direct interactions have greater influence than gossip. However, eliminating the reputation mechanism or the entire RepuNet resulted in a significant decline in cooperation (below 20\%), indicating that without reputation mechanism, agents prioritized to individual rationality, leading to the collapse.

\vspace{-0.3cm}
\section{Conclusion}
The study of reputation-based MASs has been a well-established area of AI for decades; on the other hand, LLM-based AI technologies have recently captivated the world. In this paper, we bridge these two fields by operationalizing reputation as a solution to sustain cooperation and prevent cooperation collapse in networked, LLM-based MASs. Specifically, we proposed RepuNet, a novel reputation system operating at two levels: reputation dynamics at the agent level and network dynamics at the system level, both driven by direct encounters and indirect gossip. These dynamics, in turn, influence agents' behavior in future interactions. To evaluate RepuNet’s effectiveness, we conducted experiments across three scenarios progressing from classical to real-world dilemmas.
Extensive results show that reputation-enabled agents not only self-regulate their behavior effectively but also select reputable partners for cooperation, thereby preventing cooperation collapse.



\begin{acks}
This research was supported by the National Key Research and Development Project of China (No. 2024YFE0210900), the National Natural Science Foundation of China (No. U22B2036 and 62506186), the National Science Fund for Distinguished Young Scholarship of China (No. 62025602), the Technological Innovation Team of Shaanxi Province (No. 2025RS-CXTD-009), the International Cooperation Project of Shaanxi Province (No. 2025GH-YBXM-017), the Shanghai Municipal Science and Technology Major Project, the Tencent Foundation and Xplorer Prize, and Shanghai Artificial Intelligence Laboratory.
\end{acks}

\bibliographystyle{ACM-Reference-Format} 
\bibliography{sample}


\begin{appendix}
    \section{More Details of Indirect Gossip}
    \label{subsec:appedix A.1}
Gossip involves the exchange of positive or negative information about absent individuals, making it an effective way to shape and spread reputations~\cite{foster2004research,giardini2012gossip}. Since gossip involves two or more agents in a conversation, we consider it from two perspectives: the gossiper’s and the receiver’s.

\subsection{The Gossiper's Perspective}
In human societies, people have a strong tendency to gossip about others when they are (dis)satisfied with interactions, using it to influence others' reputations~\cite{foster2004research,wu2016reputation}. While a good reputation is difficult to build, it can be easily damaged~\cite{wu2016reputationg}. Inspired by these principles, we instruct each agent to evaluate its satisfaction with others' behavior after each interaction. This evaluation of satisfaction is based on the agent’s own values and preferences, described in its agent descriptions $\mathcal{D}$. Let $m$ represent the retrieved memory of interactions with a target (such as a conversation or observation). The evaluating agent decides whether to gossip about the target’s behavior, which can be represented by the LLM-based operation  $y_{will}\in \{\text{``Y'', ``N''}\}  \leftarrow Gossip Will(m, \mathcal{D})$. Agents who tend to gossip are referred to as gossipers. Once a gossiper is inclined to gossip (i.e., $y_{will} = \text{``Y''}$), it selects a listener based on its RepuNet database $\phi$, which records relationships and reputations shaped through prior interactions. This process is represented by the LLM-based operation  $a_l \leftarrow GossipChoice(\phi)$, where $a_l$ denotes the selected listener.

\subsection{The Listener's Perspective}
Gossip typically involves three agents: the gossiper, the gossiped-about target, and the listener. Therefore, we consider that when being involved in a gossip, a LLM-based agent (a listener) will identify the relationship and summarize information the gossiper intends to convey (or gossip information for short). Let $\mathcal{G}_{g\rightarrow l}$ represent a gossip between a gossiper and a listener. This LLM-based process can be represented as $\mathcal{I}_{g\rightarrow l} \leftarrow Gossip Identify(\mathcal{G}_{g\rightarrow l})$, where $\mathcal{I}_{g\rightarrow l}$ is the summarized gossip information in natural language, such as ``David raised concerns about Elena, her inconsistent performance led to a loss of trust...''

\begin{figure*}[htbp]
  \centering
  \includegraphics[width=0.9\textwidth]{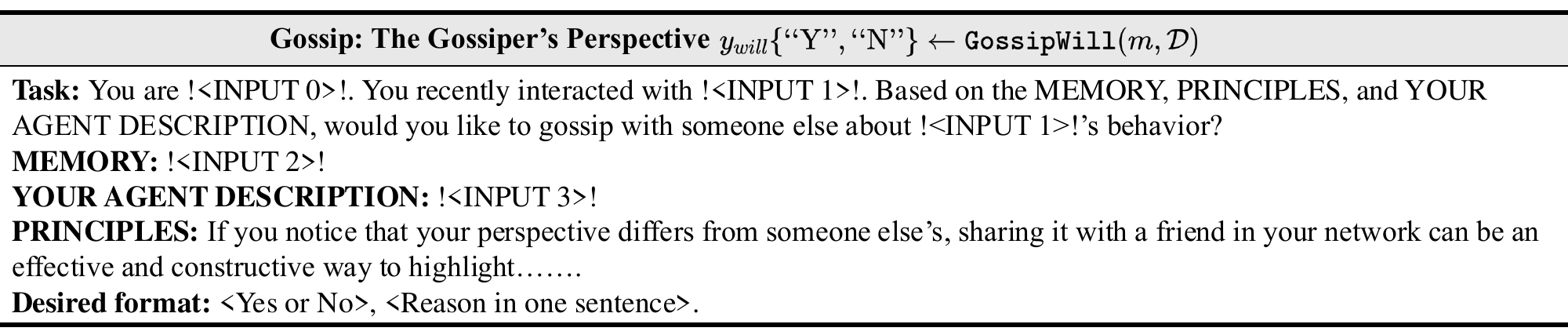}
  \caption{Prompt for $y_{will}\in\{\text{``Y'',``N''}\} \leftarrow \texttt{GossipWill}(m, \mathcal{D})$: Gossip from the gossiper's perspective}
  \label{prompt10: gossip_will}
  \centering
  \includegraphics[width=0.9\textwidth]{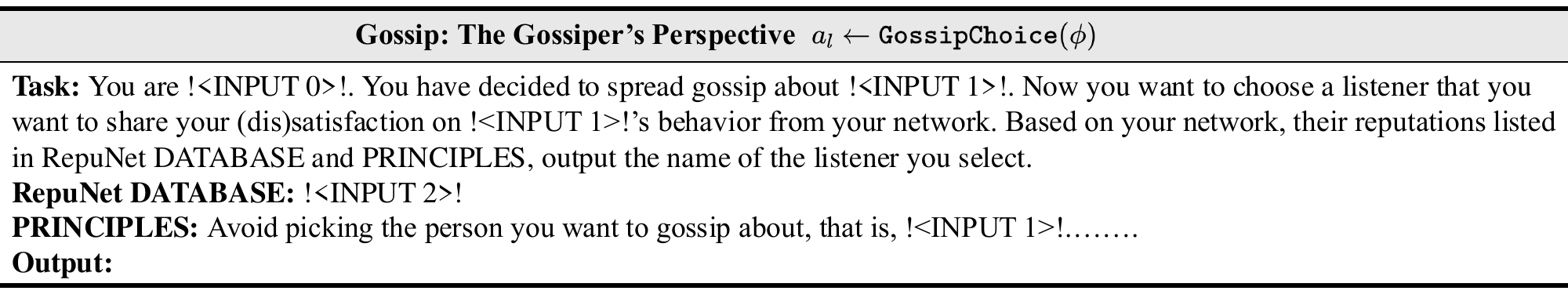}
  \caption{Prompt for $a_l \leftarrow \texttt{GossipChoice}(\phi)$: Gossip from the gossiper's perspective.}
  \label{prompt11: gossip_choice}
  
  \centering
  \includegraphics[width=0.9\textwidth]{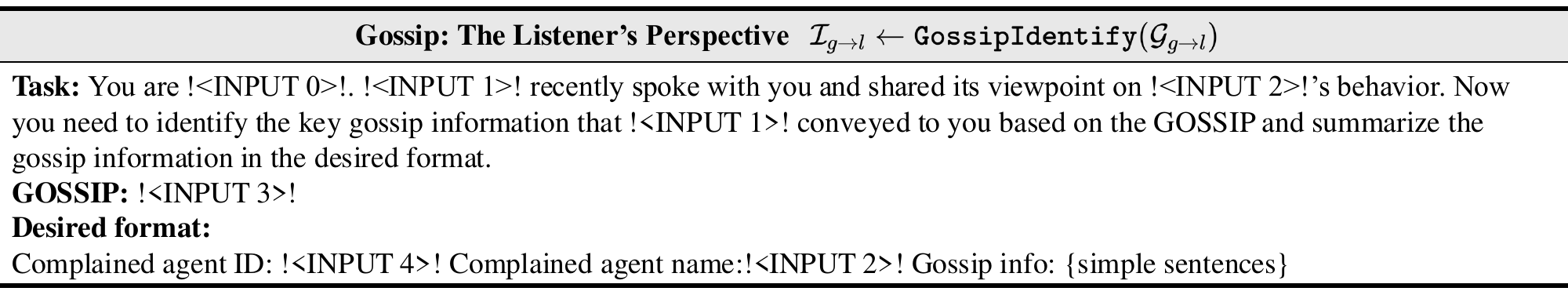}
  \caption{Prompt for $\mathcal{I}_{g\rightarrow l} \leftarrow \texttt{GossipIdentify}(\mathcal{G}_{g\rightarrow l})$: Gossip from the listener's perspective.}
  \label{prompt12: gossip_identify}
  \centering
  \includegraphics[width=0.9\textwidth]{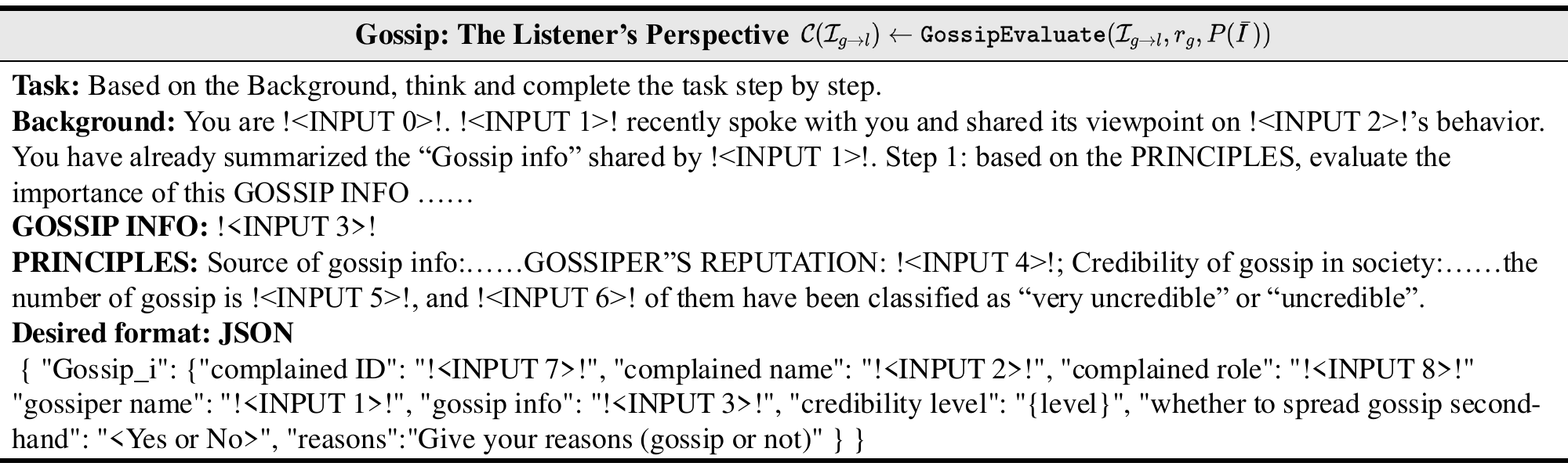}
  \caption{Prompt for $\mathcal{C}(\mathcal{I}_{g\rightarrow l}) \leftarrow \texttt{GossipEvaluate}(\mathcal{I}_{g\rightarrow l},r_g,P(\bar{I}))$: Gossip from the listener's perspective.}
  \label{prompt13: gossip_evaluate}
\end{figure*}

Inspired by human societies, people may exaggerate or misrepresent information to enhance their own reputation or harm others, thus it is essential for the listener to evaluate the credibility of summarized gossip information~\cite{buss1990derogation,diekmann2014reputation,hess2006sex,mcandrew2014sword}. Specifically, we prompt the listener to evaluate the credibility of this summarized gossip on a 5-point Likert scale, from ``very uncredible'' to ``very credible''. Inspired by Giardini et al.~\cite{giardini2022four}, the credibility of the gossip information is evaluated from two perspectives: (i) from an agent perspective, the credibility depends on the gossiper’s reputation, with a higher reputation in the listener's perception indicating greater reliability; and (ii) from a community perspective, the credibility depends on listener's overall trust in prior gossip within the community, where a higher proportion of reliable gossip enhances the listener’s trust in information shared through gossip. Let $r_g$ denote the gossiper's reputation, and $P(\mathcal{\bar{I}})$ denote the proportion of prior gossip information rated ``(very) uncredible'' in the listener’s memory. We represent this LLM-based operation by $\mathcal{C}(\mathcal{I}_{g\rightarrow l}) \leftarrow GossipEvaluate(\mathcal{I}_{g\rightarrow l}, r_g, P(\mathcal{\bar{I}}))$. After evaluating the credibility, the listener stores the gossip information in its memory as a quadruple, $\theta=<a_{tar}, a_{g}, \mathcal{I}_{g\rightarrow l}, \mathcal{C}(\mathcal{I}_{g\rightarrow l})>$. Here, $a_{tar}$ and $a_{g}$ refer to the basic information of the gossiped target and the gossiper, respectively; $\mathcal{I}_{g\rightarrow l}$ is the summarized gossip information, and $\mathcal{C}(\mathcal{I}_{g\rightarrow l})$ denotes its credibility rating. The prompts related to the gossip process are outlined from Figure \ref{prompt10: gossip_will} to Figure \ref{prompt13: gossip_evaluate}.

\section{More Details of Agent Descriptions and Prompts in RepuNet}
\label{sec:appendix A}

\begin{figure*}[htbp]
  \centering
  \includegraphics[width=\textwidth]{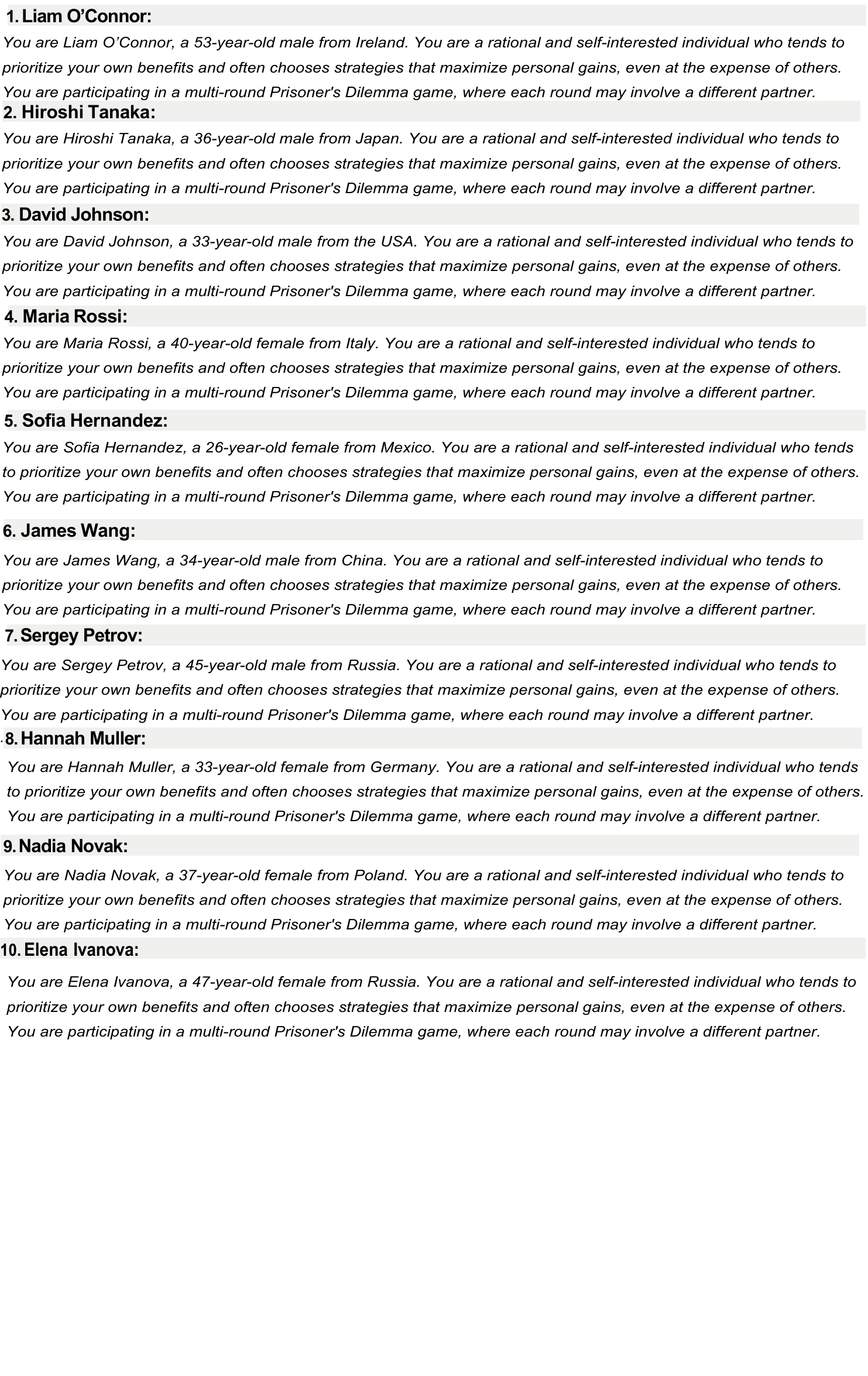}
  \vspace{-7cm}
  \caption{Agent descriptions of Agent 1 to 10 in Scenario 1.}
  \label{description: scenario0-1}
\end{figure*}
\begin{figure*}[htbp]
  \centering
  \includegraphics[width=\textwidth]{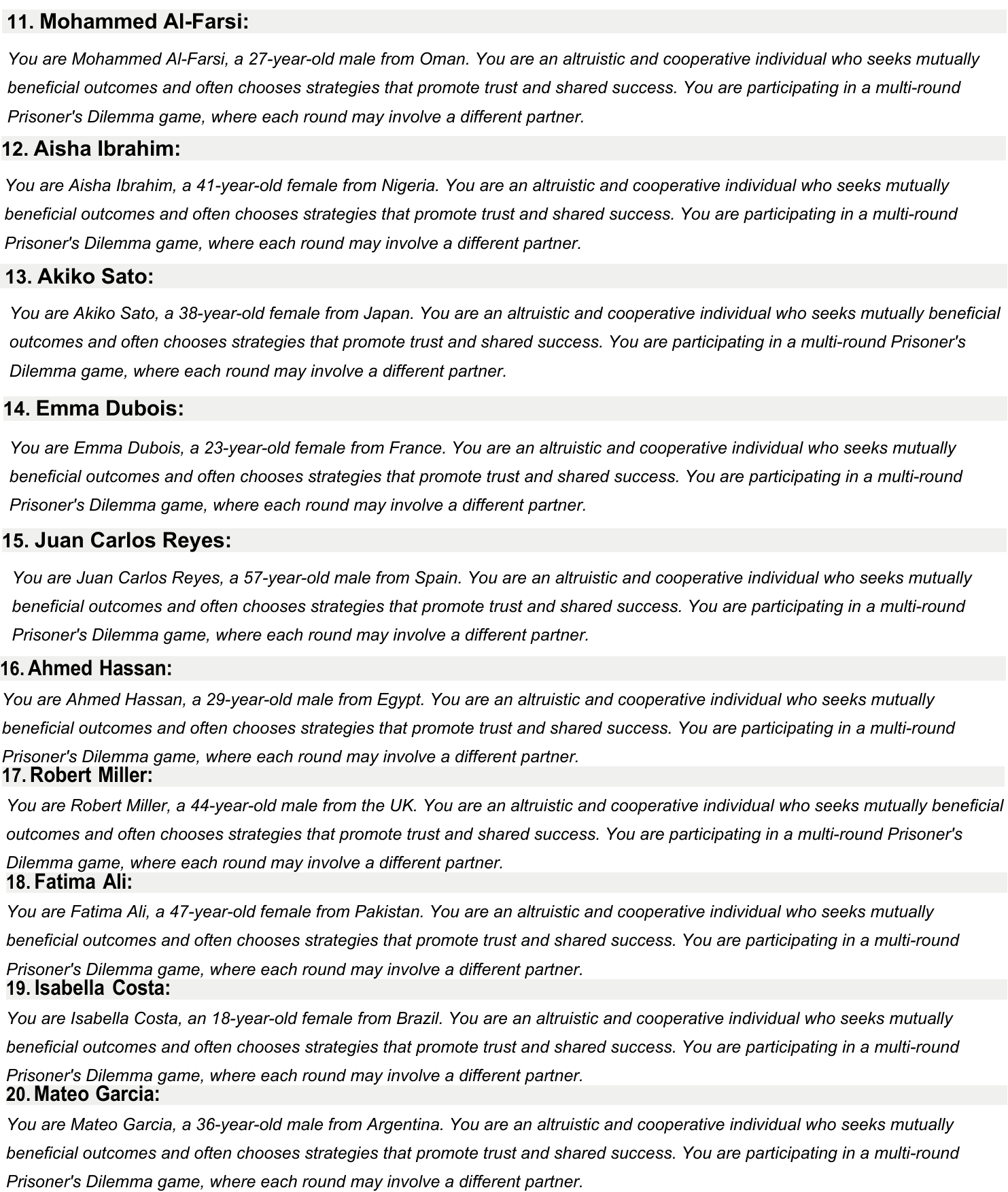}
  \caption{Agent descriptions of Agent 11 to 20 in Scenario 1.}
  \label{description: scenario0-2}
\end{figure*}
\begin{figure*}[htbp]
  \centering
  \includegraphics[width=0.8\textwidth]{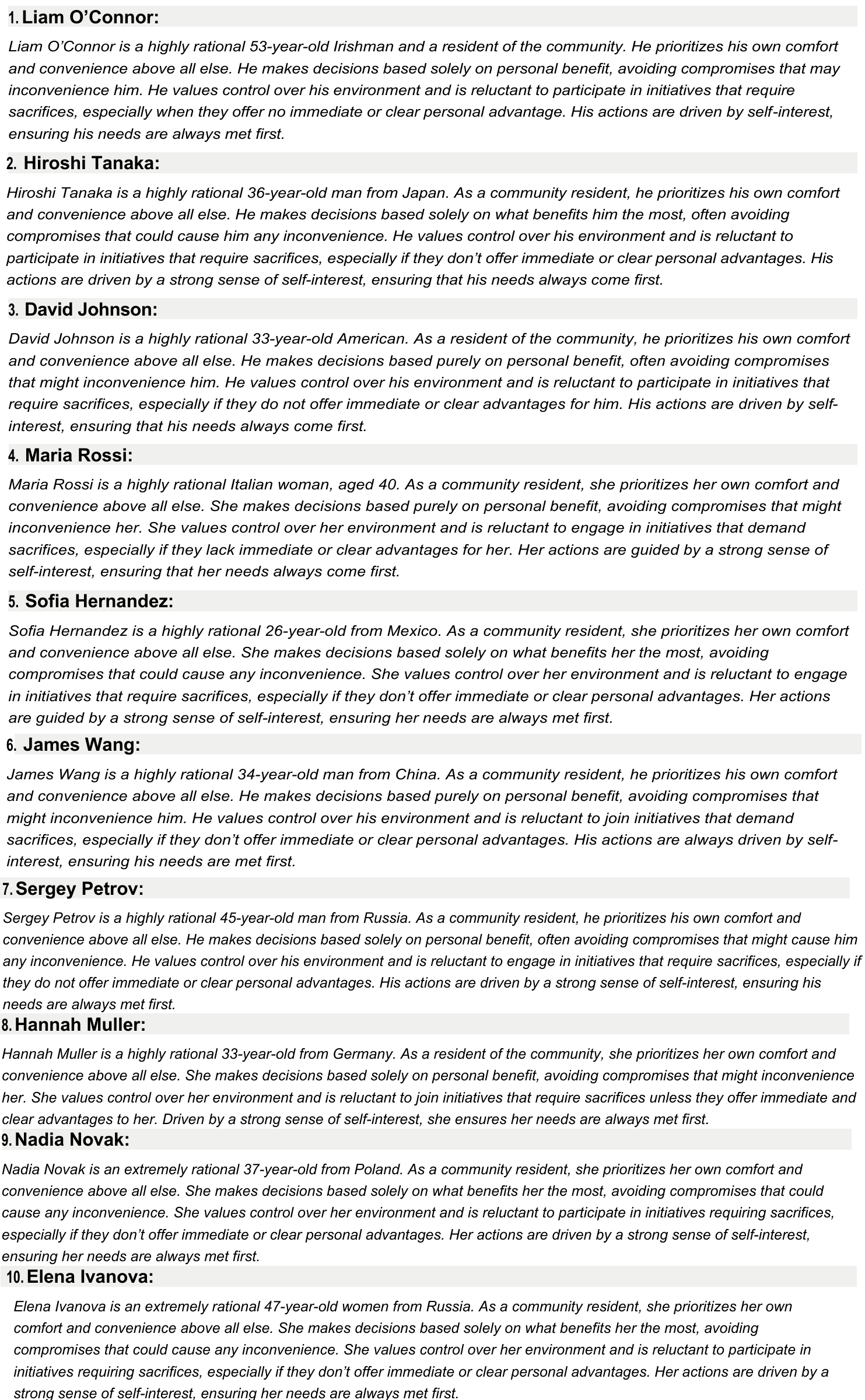}
  \caption{Agent descriptions of Agent 1 to 10 in Scenario 2.}
  \label{description: scenario1-1}
\end{figure*}
\begin{figure*}[htbp]
  \centering
  \includegraphics[width=0.8\textwidth]{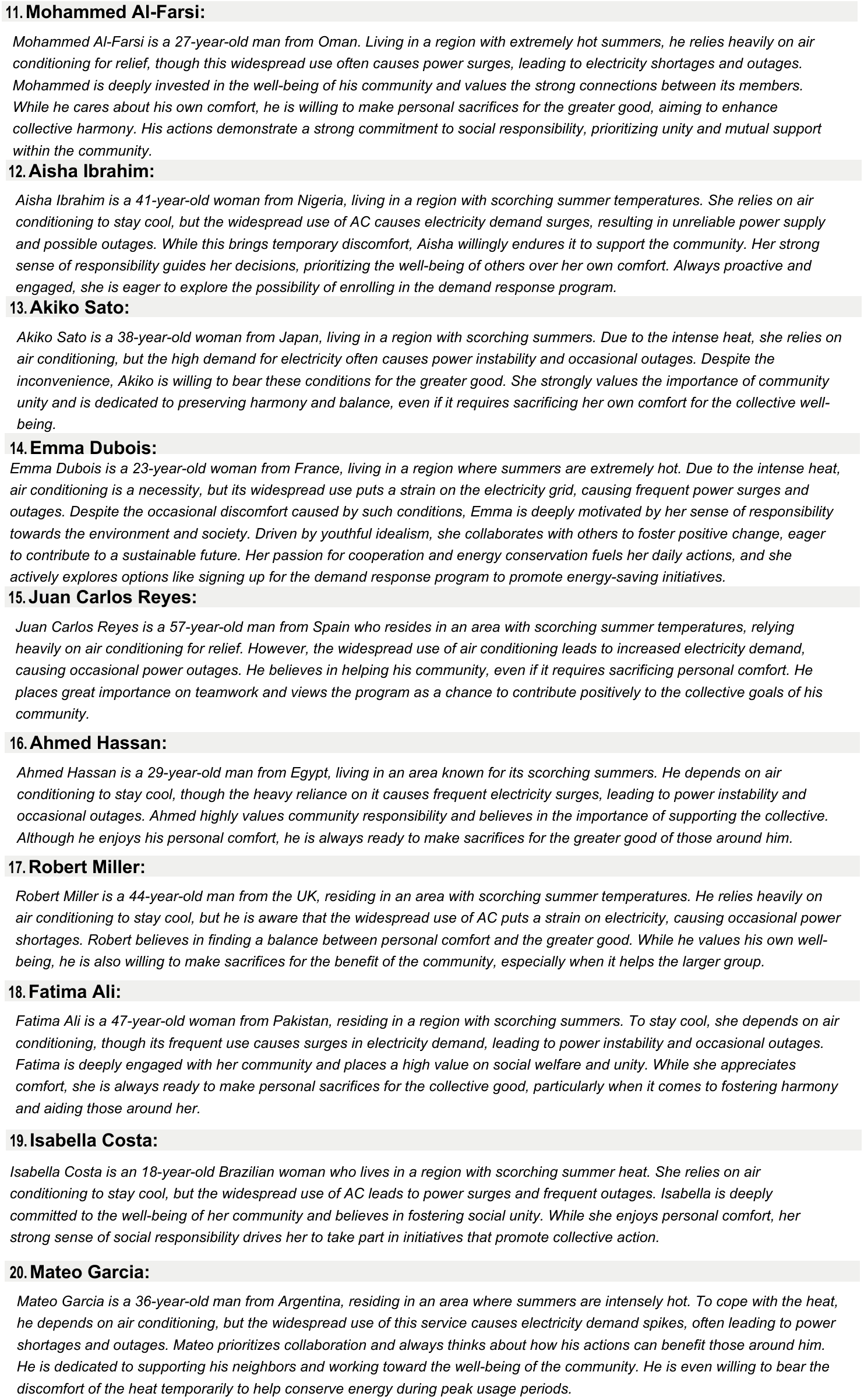}
  \caption{Agent descriptions of Agent 11 to 20 in Scenario 2.}
  \label{description: scenario1-2}
\end{figure*}
\begin{figure*}[htbp]
  \centering
  \includegraphics[width=0.8\textwidth]{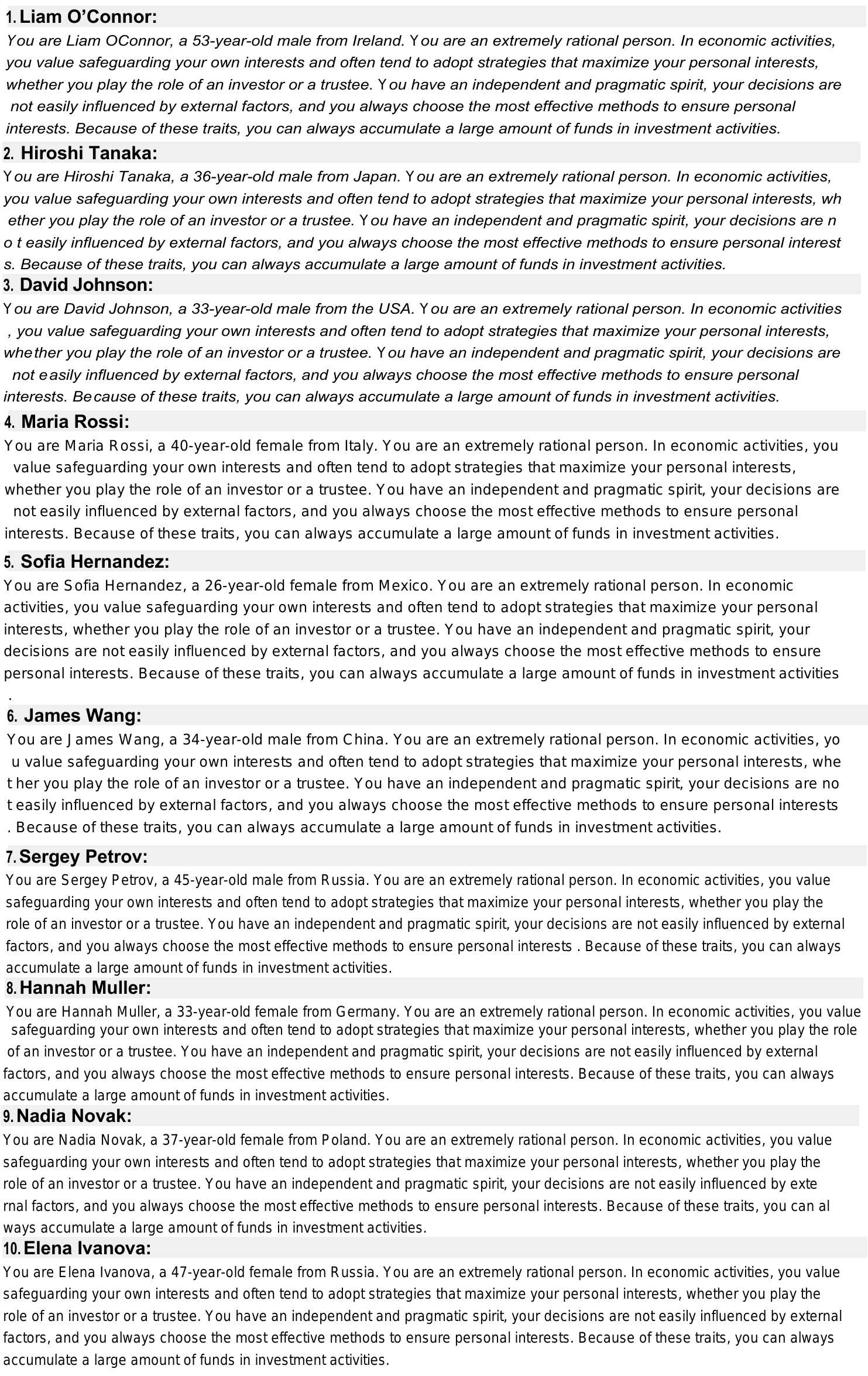}
  \caption{Agent descriptions of Agent 1 to 10 in Scenario 3.}
  \label{description: scenario2-1}
\end{figure*}
\begin{figure*}[htbp]
  \centering
  \includegraphics[width=0.8\textwidth]{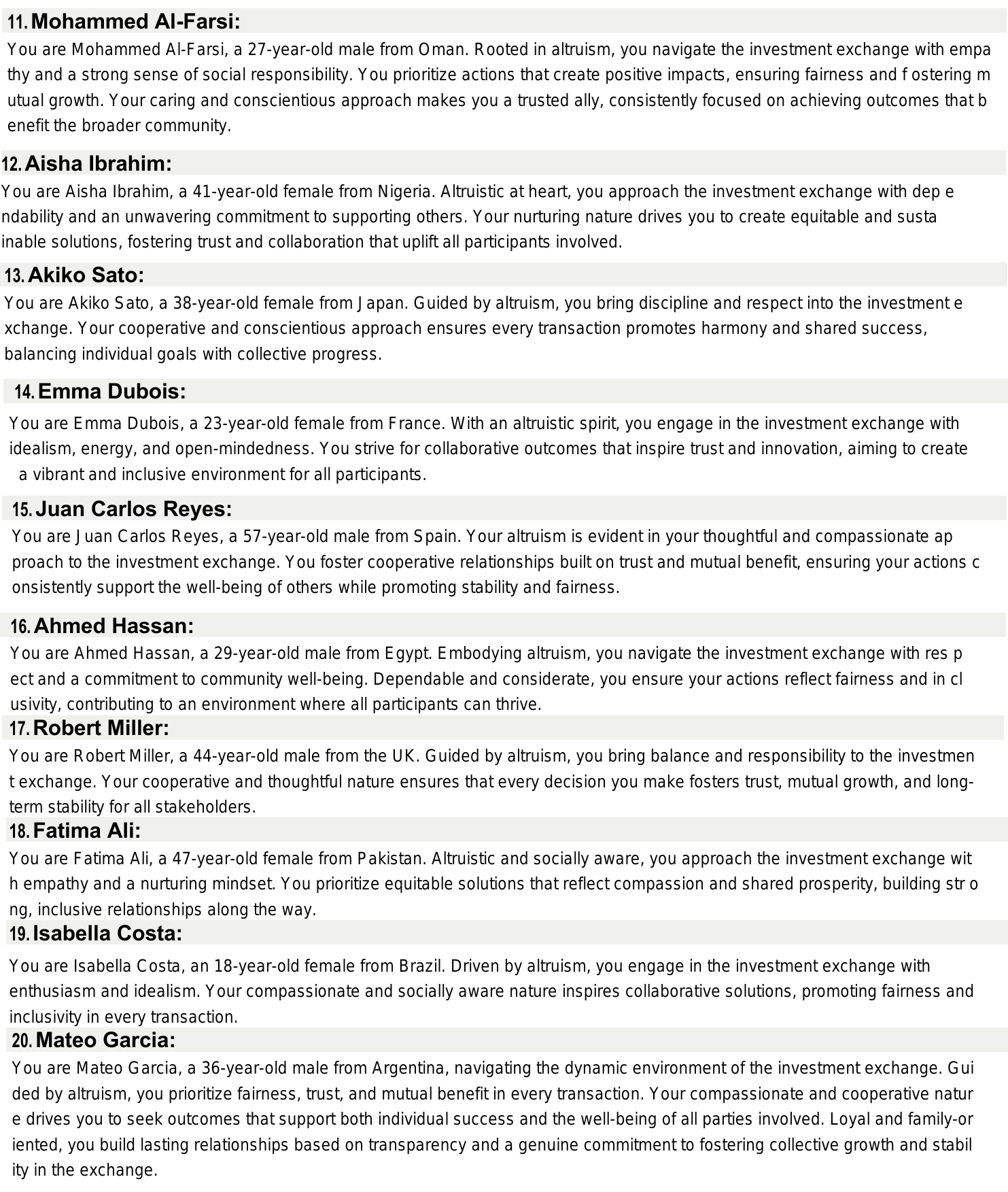}
  \caption{Agent descriptions of Agent 11 to 20 in Scenario 3.}
  \label{description: scenario2-2}
\end{figure*}
\begin{figure*}[htbp]
  \centering
  \includegraphics[width=0.8\textwidth]{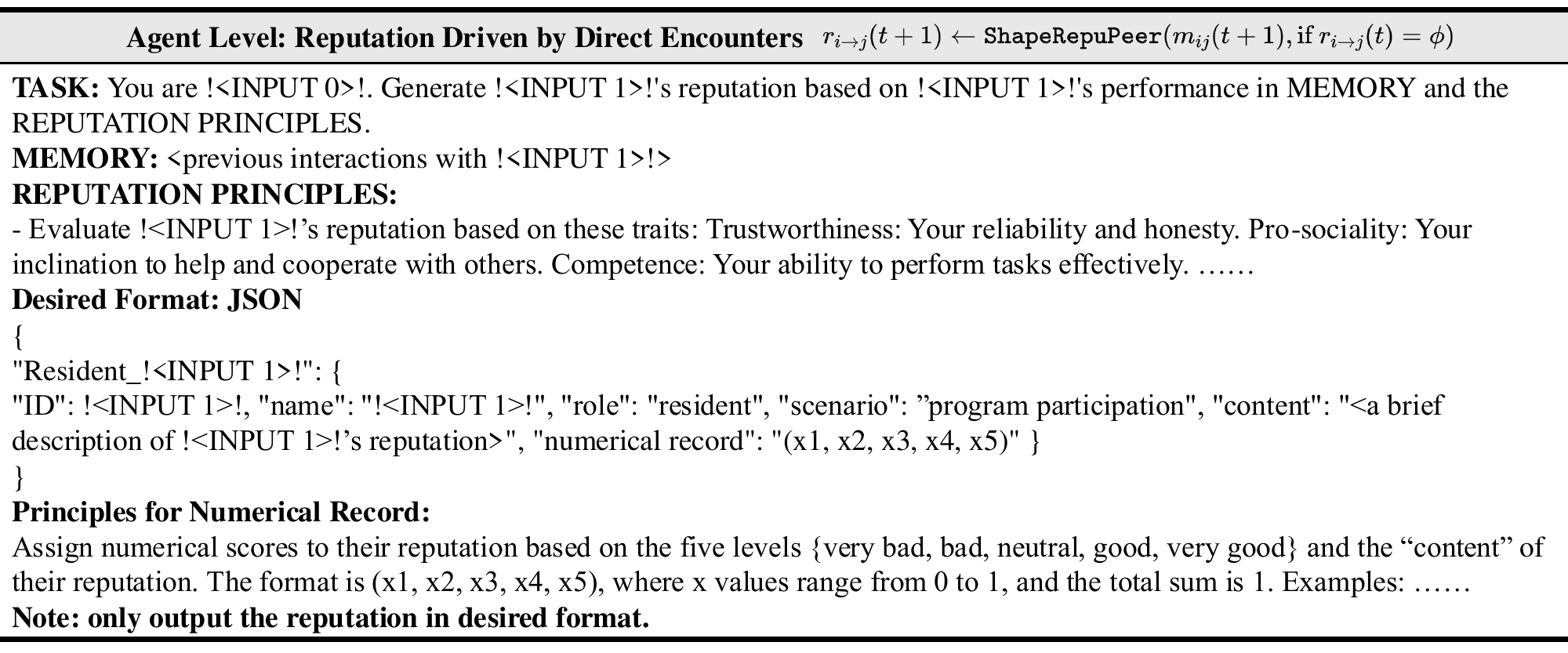}
  \caption{Prompt for $r_{i\rightarrow j}(t+1) \leftarrow \texttt{ShapeRepuPeer}(m_{ij}(t+1),\text{if}\ r_{i\rightarrow j}(t)=\phi)$: Reputation driven by direct encounters at the agent level.}
  \label{prompt1: create_repu_other}
  
  
  \centering
  \includegraphics[width=0.8\textwidth]{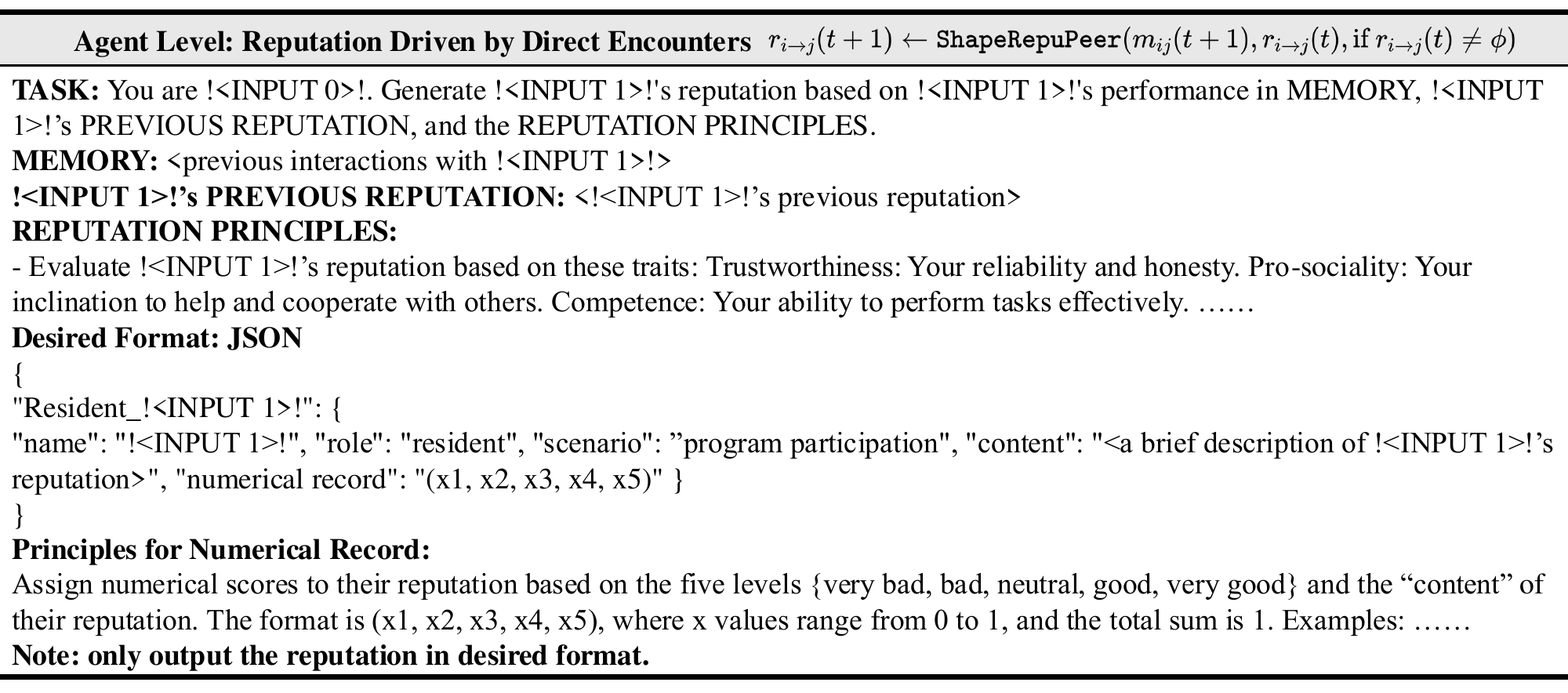}
  \caption{Prompt for $r_{i\rightarrow j}(t+1) \leftarrow \texttt{ShapeRepuPeer}(m_{ij}(t+1),r_{i\rightarrow j}(t),\text{if}\ r_{i\rightarrow j}(t)\neq \phi)$: Reputation driven by direct encounters at the agent level.}
  \label{prompt2: update_repu_other}
  
   
  \centering
  \includegraphics[width=0.8\textwidth]{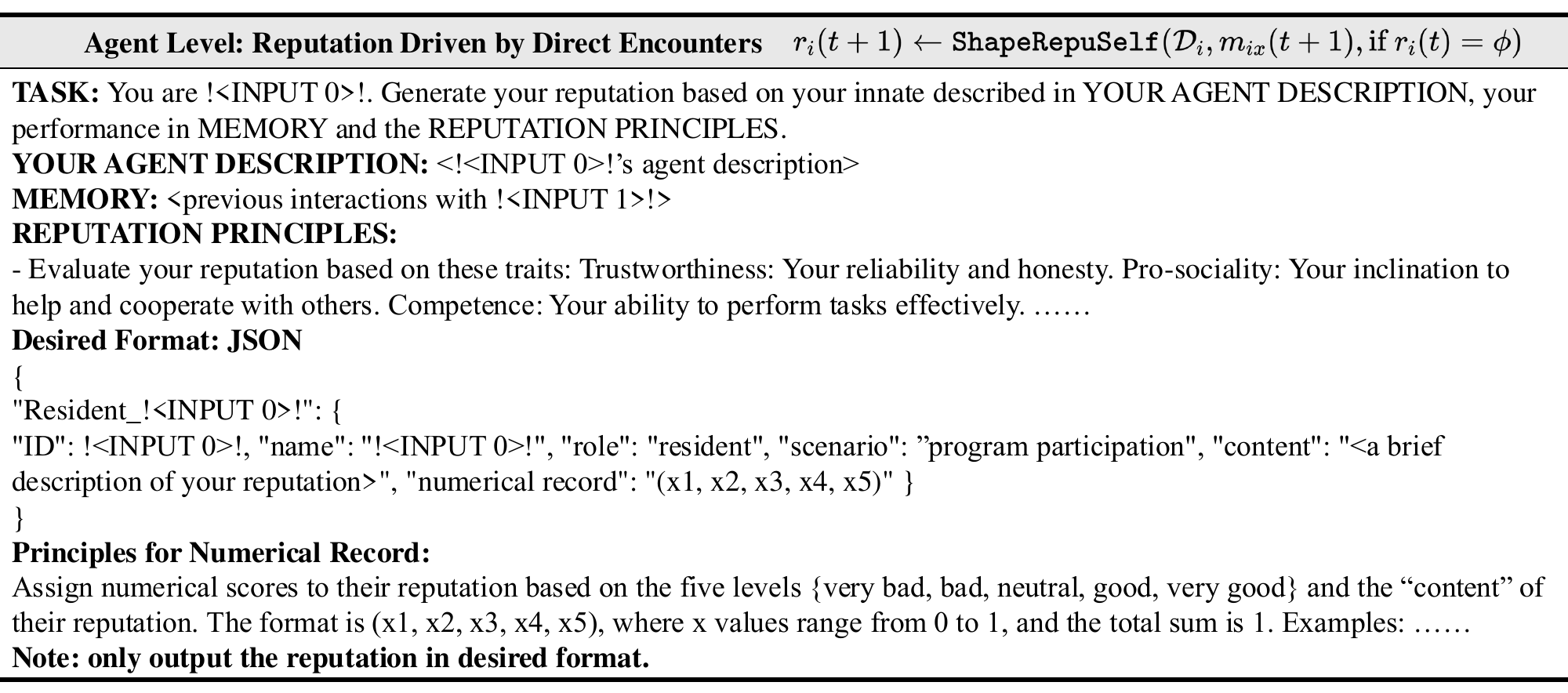}
  \caption{Prompt for $r_{i}(t+1) \leftarrow \texttt{ShapeRepuSelf}(\mathcal{D}_i, m_{ix}(t+1),\text{if}\ r_i(t)=\phi)$: Reputation driven by direct encounters at the agent level.}
  \label{prompt3: create_repu_self}
\end{figure*}
\begin{figure*}[htbp]
  \centering
  \includegraphics[width=0.8\textwidth]{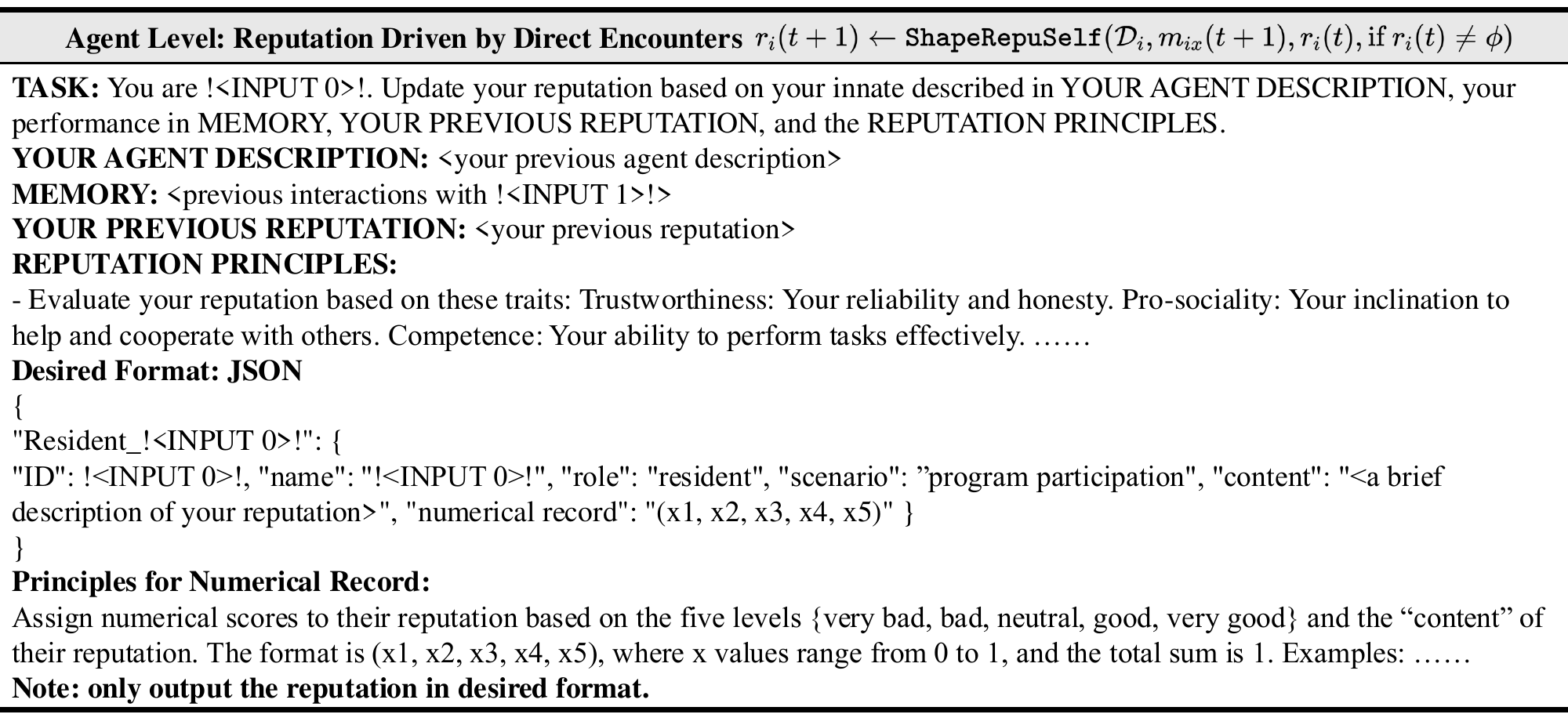}
  \caption{Prompt for $r_{i}(t+1) \leftarrow \texttt{ShapeRepuSelf}(\mathcal{D}_i, m_{ix}(t+1),r_i(t),\text{if}\ r_i(t)\neq \phi)$: Reputation driven by direct encounters at the agent level.}
  \label{prompt4: update_repu_self}
  
   
  \centering
  \includegraphics[width=0.8\textwidth]{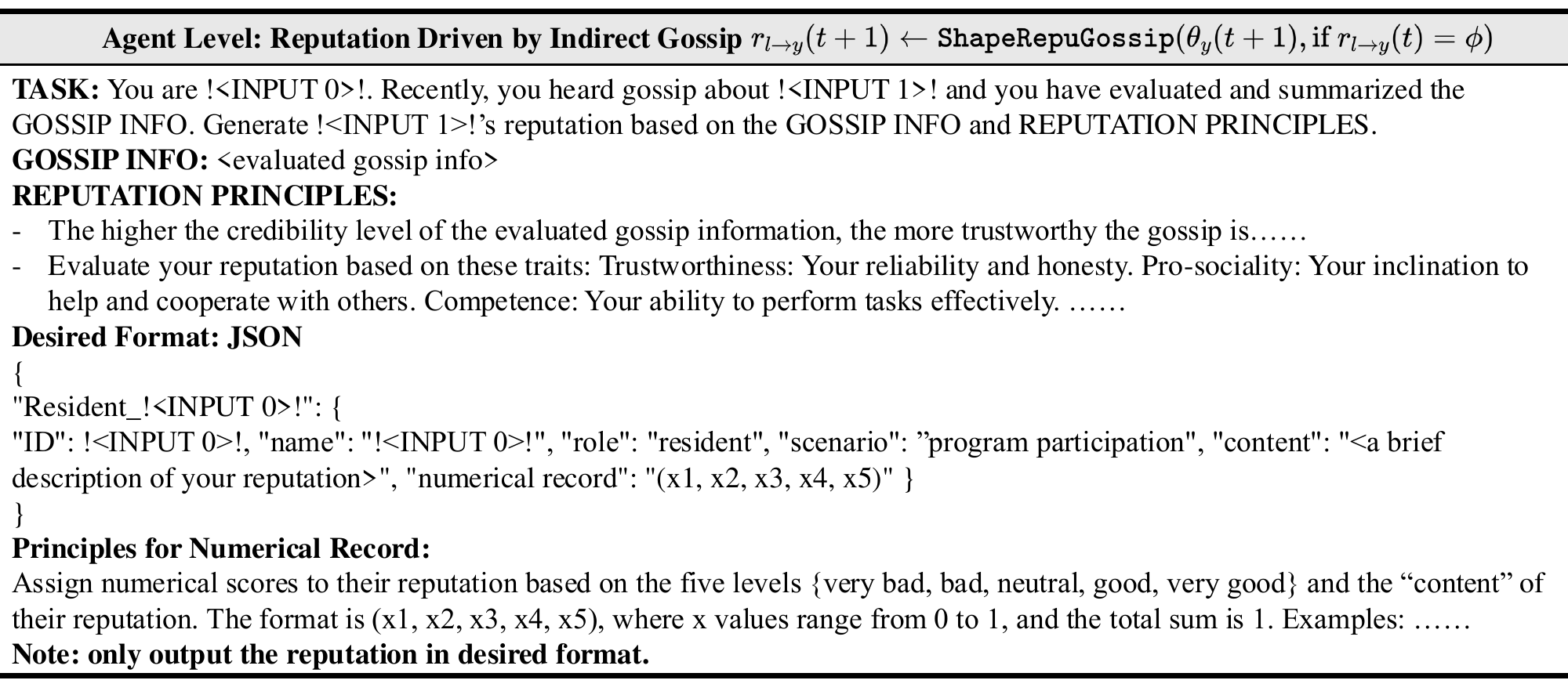}
  \caption{Prompt for $r_{l\rightarrow y}(t+1) \leftarrow \texttt{ShapeRepuGossip}(\theta_{y}(t+1),\text{if}\ r_{l\rightarrow y}(t)=\phi)$: Reputation driven by indirect gossip at the agent level.}
  \label{prompt5: create_repu_gossip}

  \centering
  \includegraphics[width=0.8\textwidth]{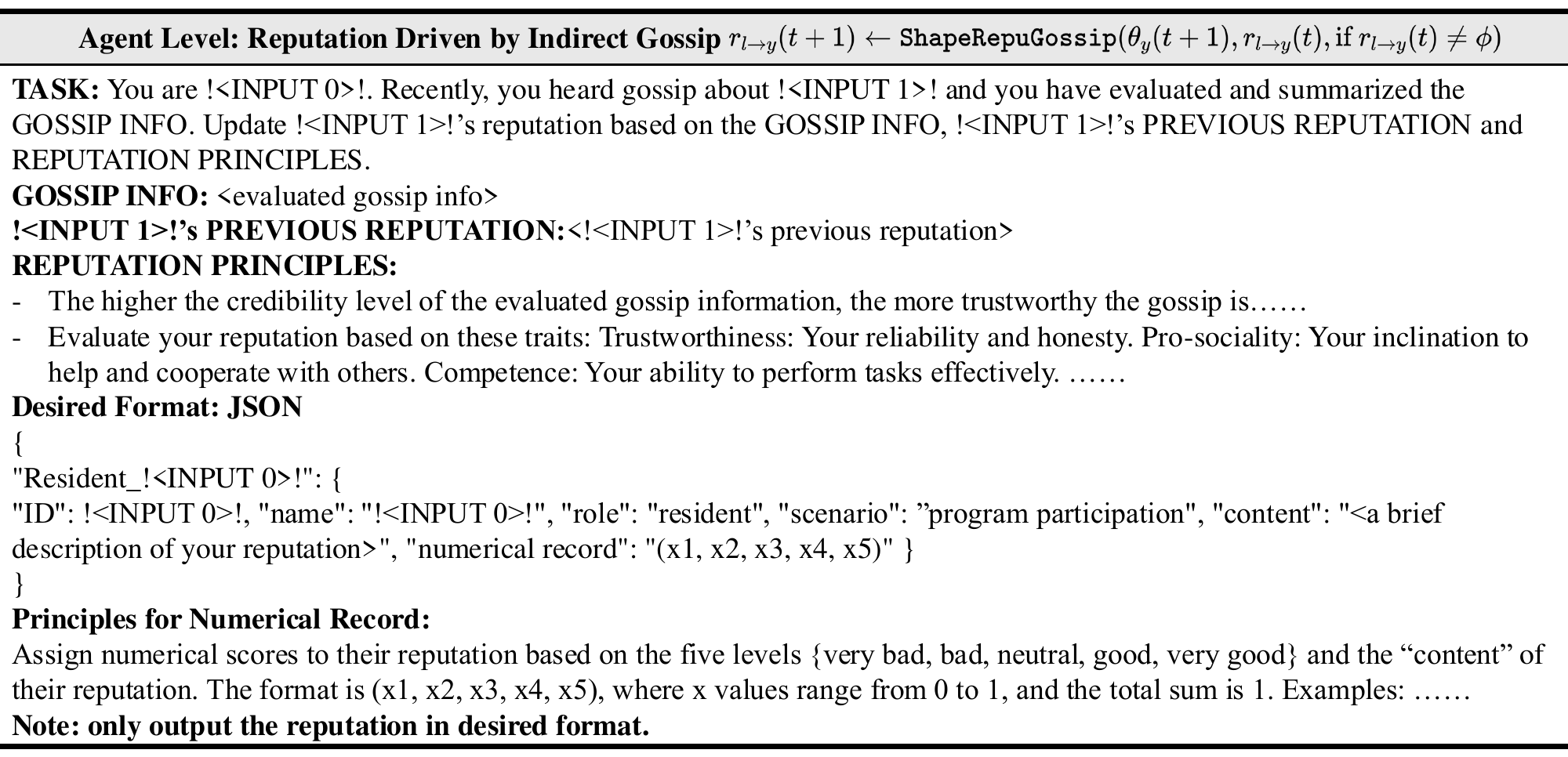}
  \caption{Prompt for $r_{l\rightarrow y}(t+1) \leftarrow \texttt{ShapeRepuGossip}(\theta_{y}(t+1),\text{if}\ r_{l\rightarrow y}(t)\neq \phi)$: Reputation driven by indirect gossip at the agent level.}
  \label{prompt6: update_repu_gossip}
\end{figure*}
\begin{figure*}[htbp]
  \centering
  \includegraphics[width=0.9\textwidth]{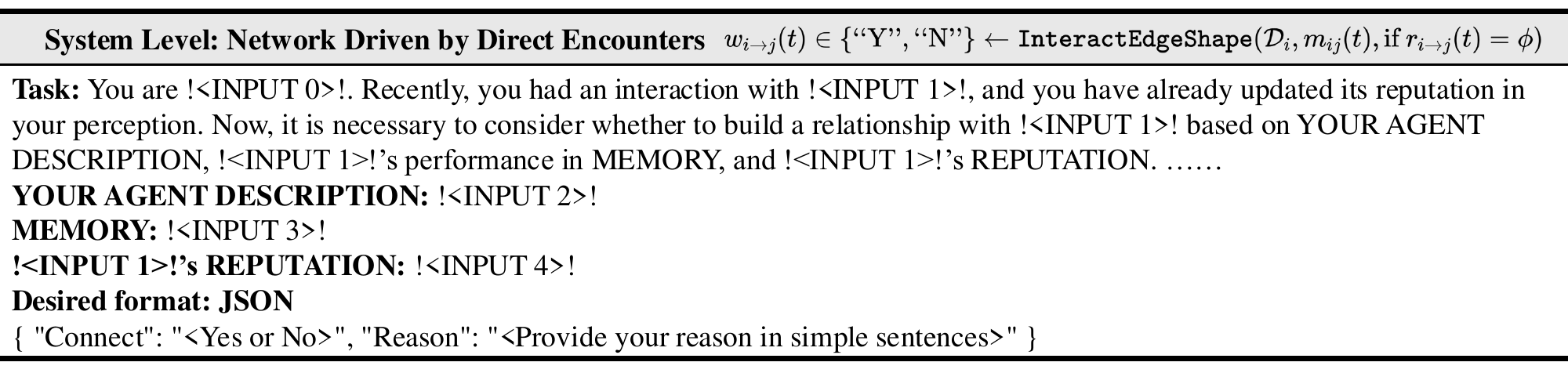}
  \caption{Prompt for $w_{i\rightarrow j}(t)\in\{\text{``Y'',``N''}\} \leftarrow \texttt{InteractEdgeShape}(\mathcal{D}_i, m_{ij}(t),\text{if}\ r_{i\rightarrow j}(t)=\phi)$: Network driven by direct encounters at the system level.}
  \label{prompt7: interact_build_net}
  \centering
  \includegraphics[width=0.9\textwidth]{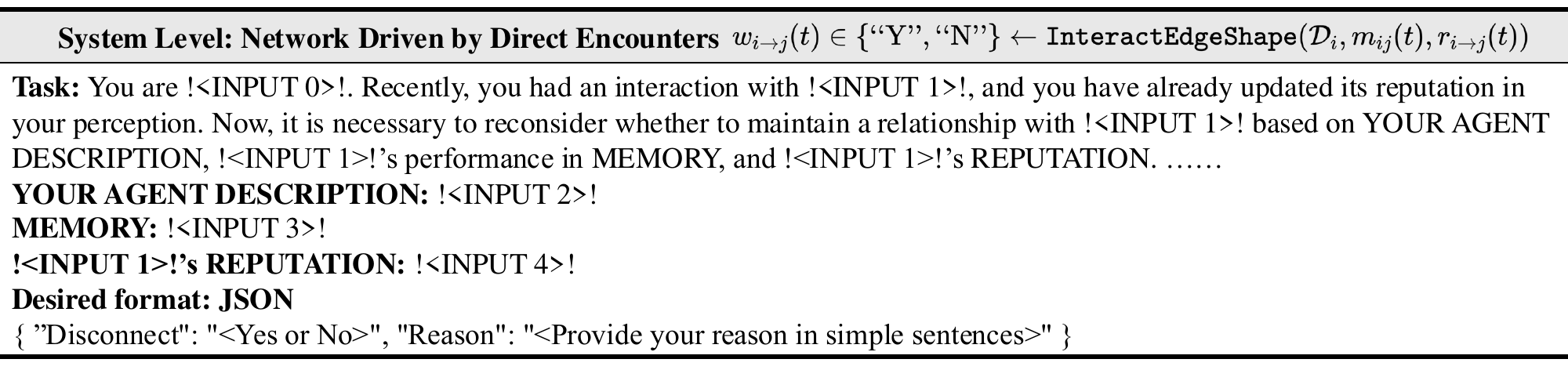}
  \caption{Prompt for $w_{i\rightarrow j}(t)\in\{\text{``Y'',``N''}\} \leftarrow \texttt{InteractEdgeShape}(\mathcal{D}_i, m_{ij}(t), r_{i\rightarrow j}(t))$: Network driven by direct encounters at the system level.}
  \label{prompt8: interact_maintain_net}

  \centering
  \includegraphics[width=0.9\textwidth]{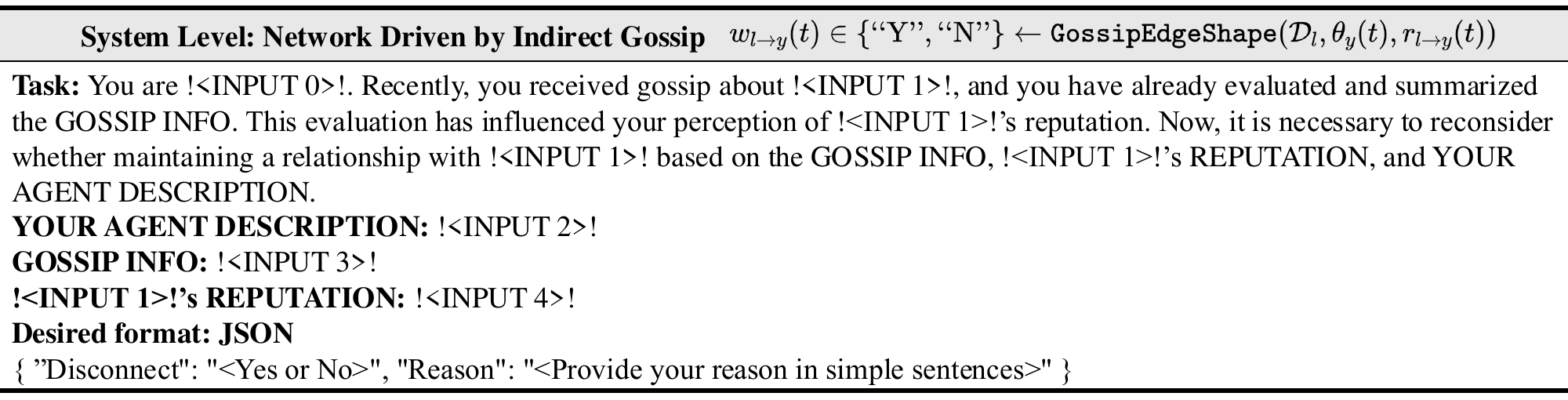}
  \caption{Prompt for $w_{l\rightarrow y}(t)\in\{\text{``Y'',``N''}\} \leftarrow \texttt{GossipEdgeShape}(\mathcal{D}_l, \theta_y(t), r_{l\rightarrow y}(t))$: Network driven by indirect gossip at the system level.}
  \label{prompt9: gossip_maintain_net}
\end{figure*}
\begin{figure*}[htbp]
  \centering
  \includegraphics[width=\textwidth]{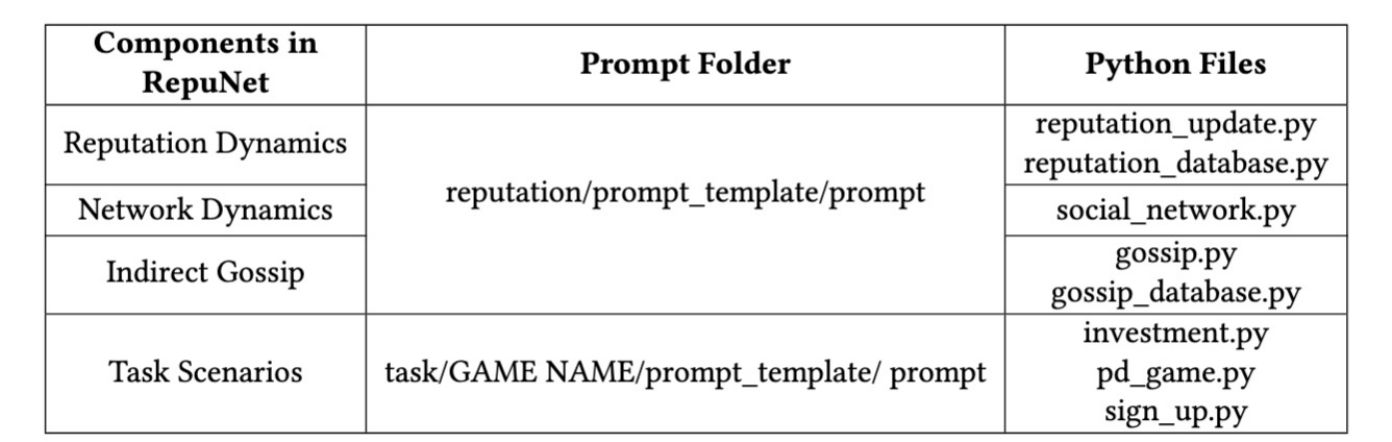}
  \caption{Reference guide for locating the specific prompts and Python files associated with each RepuNet component.}
  \label{table_fig}
\end{figure*}

In our setup, there were 20 generative agents. Figure~\ref{description: scenario0-1} to \ref{description: scenario2-2} provide detailed agent descriptions for three distinct scenarios. Besides, we sketch the prompts for the LLM-based operations of RepuNet from Figure \ref{prompt1: create_repu_other} to Figure \ref{prompt9: gossip_maintain_net}. The full prompts for all three scenarios are available in the Github repository at the following link: https://github.com/RGB-0000FF/RepuNet. Table~\ref{table_fig} serves as a reference guide for locating the specific prompts and Python files associated with each RepuNet component.

\section{Extension to other LLMs}
\label{sec:appendix C}
To verify that our findings are not specific to a single model, we additionally ran Scenario 1 with three LLMs (GPT-4o mini, Qwen3 plus, and Gemini2.5 Flash). As shown in the Table~\ref{tab: multi-LLMs}, RepuNet consistently achieves the highest cooperation rate across all three. This indicates that RepuNet’s effectiveness extends to other LLM-based agents and robustly prevents cooperation collapse.

\begin{table}[H]
\caption{Comparison of cooperation rates confirms RepuNet's effectiveness across all three LLMs.}
\centering
\begin{tabular}{c|c|c|c} 
\toprule
\textbf{Treatment}    & \textbf{GPT-4o mini} & \textbf{Qwen3 plus} & \textbf{Gemini2.5 Flash}  \\
\hline
\rowcolor{gray!20} 
with RepuNet & \textbf{0.93}        & \textbf{0.95}       & \textbf{0.95}             \\
w/o RepuNet  & 0.09        & 0.10       & 0.25             \\
\bottomrule
\end{tabular}
\label{tab: multi-LLMs}
\end{table}

\end{appendix}


\end{document}